\documentclass{article}
\usepackage[final]{neurips_2024}
\makeatletter
\let\NAT@orig@cite\cite
\renewcommand{\cite}{\@ifnextchar[{\citep}{\citep}}
\makeatother

\usepackage{amsmath,amsfonts,bm}

\def\eqref#1{equation~\ref{#1}}

\def\1{\bm{1}}

\DeclareMathAlphabet{\mathsfit}{\encodingdefault}{\sfdefault}{m}{sl}
\SetMathAlphabet{\mathsfit}{bold}{\encodingdefault}{\sfdefault}{bx}{n}

\usepackage{url}
\usepackage{graphicx}
\usepackage{mathtools}
\usepackage{amsthm}
\usepackage{bm}

\usepackage{booktabs}
\usepackage{multirow}
\usepackage{array}
\usepackage{colortbl}
\usepackage{longtable}
\usepackage{wrapfig}
\usepackage{caption}        
\usepackage{subfigure}      

\usepackage[table,dvipsnames]{xcolor}

\usepackage{enumitem}
\usepackage{float}
\usepackage{rotating}

\usepackage{pifont}
\usepackage{adjustbox}
\usepackage{siunitx}
\usepackage{tablefootnote}
\usepackage{authblk}

\usepackage[most]{tcolorbox}
\usepackage[ruled,vlined,linesnumbered]{algorithm2e}
\DontPrintSemicolon
\usepackage{todonotes}

\usepackage{etoc}
\usepackage{minitoc}

\usepackage{orcidlink}

\usepackage[capitalize]{cleveref}

\newcommand{\grayrow}{\rowcolor[gray]{.95}}



\definecolor{mybackblue}{RGB}{240,248,255}
\definecolor{myframeblue}{RGB}{100,149,237}
\newtcolorbox{keyfindingbox}[1][]{%
  enhanced, breakable,
  colback=mybackblue, colframe=myframeblue,
  coltitle=white, fonttitle=\bfseries\small,
  attach boxed title to top left={yshift=-2mm, xshift=2mm},
  boxed title style={colback=myframeblue, sharp corners=south, boxrule=0pt, arc=3pt,
    left=6pt,right=6pt,top=2pt,bottom=2pt},
  title=#1,
  boxrule=0.4pt, arc=2mm,
  left=8pt,right=8pt,top=8pt,bottom=8pt
}

\definecolor{lightred}{HTML}{e54c3e}
\definecolor{lightyellow}{HTML}{f2b705}
\definecolor{lightblue}{HTML}{3498db}

\begin{document}
\etocdepthtag.toc{mtchapter}
\etocsettagdepth{mtchapter}{none}
\etocsettagdepth{mtappendix}{none}
\title{AesRM: Improving Video Aesthetics with Expert-Level Feedback} 
\author{%
Yujin Han$^{1,6}$\thanks{Equal contribution. Email to: Yujin Han (yujinhan@connect.hku.hk).},
Yujie Wei$^{2,*}$,
Yefei He$^{3}$,
Xinyu Liu$^{4}$,
Tianle Li$^{1}$,
Zichao Yu$^{1}$,\\
Andi Han$^{5}$,
Shiwei Zhang$^{6}$,
Tingyu Weng$^{6,\dagger}$,
Difan Zou$^{1}$\thanks{Correspondence to: Tingyu Weng and Difan Zou.}
}

\affil{\small{$^{1}$The University of Hong Kong, $^{2}$Fudan University,
$^{3}$Zhejiang University, $^{4}$Hong Kong University of Science and Technology,
$^{5}$University of Sydney, $^{6}$Alibaba Group}}


\maketitle
\vspace{-3.5em}
\begin{center}
\texttt{Project Page: \url{https://yujinhanml.github.io/aesrm-projectpage}}
\end{center}
\begin{abstract}

Despite rapid advances in photorealistic video generation, real-world applications such as filmmaking require video aesthetics, e.g., harmonious colors and cinematic lighting, beyond visual fidelity. Prior work on visual aesthetics largely focuses on images, often reducing aesthetics to coarse definitions, e.g., visual pleasure, without a rigorous and systematic evaluation. To improve video aesthetics, we propose a hierarchical rubric that decomposes video aesthetics into three core dimensions, Visual Aesthetics (VA), Visual Fidelity (VF), and Visual Plausibility (VP), with 15 fine-grained criteria, e.g., shot composition. This framework enables a large-scale expert-annotated preference dataset and an evaluation benchmark, \textbf{AesVideo-Bench}, containing about 2500 video pairs with expert annotations on VA, VF, and VP. We then build a family of Video Aesthetic Reward Models (AesRM): \textbf{AesRM-Base}, which directly predicts pairwise preferences on these dimensions to provide efficient post-training rewards and \textbf{AesRM-CoT}, which additionally generates CoT aligned with all 15 criteria to improve assessment interpretability. Specifically, we train AesRM with a three-stage progressive scheme: (1) Atomic Aesthetic Capability Learning, which strengthens AesRM’s recognition of fundamental aesthetic concepts, e.g., accurately identifying centered composition; (2) Cold-Start, aligning the model with structured reasoning protocols; and (3) GRPO, further improving evaluation accuracy. To enhance AesRM-CoT, we additionally propose self-consistency–based CoT synthesis to improve CoT quality and design CoT-based process rewards during GRPO. Extensive experiments show AesRM outperforms baselines on multiple aesthetics benchmarks and is more robust, with lower position bias. Finally, we align Wan2.2 with AesRM and observe clear \textbf{aesthetic gains} over existing aesthetic reward models. Quantitative and qualitative results illustrate AesRM enables models to generate videos with better color, more sophisticated lighting and richer details.

\end{abstract}


\section{Introduction}
\label{sec:intro}

Recently, rapid advances in generative models \cite{wan2025wan,gao2025seedance,kong2024hunyuanvideo,seedance2025seedance,lin2024open,polyak2024movie,liu2025vfx,wei2024dreamvideo,wei2024dreamvideo2,wei2025dreamrelation,liu2025revise,wei2025routing,liu2024hiprompt} have made photorealistic and controllable video generation possible.  However, high fidelity alone is insufficient for downstream applications such as film production \cite{wu2025moviebench,xiao2025captain,wu2025automated}, which demand specific video aesthetics, e.g., distinctive color grading and stylized shot composition. Despite substantial progress in image aesthetics assessment (IAA) \cite{cao2025artimuse,yi2023towards,cao2025unipercept,anwar2021survey,khurana2025measuring}, research on video aesthetics remains relatively limited \cite{qiao2025vadb}. Moreover, existing research on video aesthetics often adopts coarse aesthetic labels. For example, some methods label videos as aesthetically pleasing \cite{asarkar2019detecting,liu2025improving} and train models to predict an overall aesthetic score \cite{cao2025artimuse,schuhmann2022improved}. Nevertheless, such rewards lack interpretability and cannot pinpoint fine-grained deficiencies in specific aesthetic aspects, e.g., composition, color, or lighting, making it difficult to provide fine-grained supervision for training aesthetics-aware models.


To address this, we propose a systematic pipeline: defining multi-dimensional aesthetic criteria, curating expert-annotated data based on aesthetic definition and building interpretable reward models to provide granular signals for video aesthetic alignment. Specifically, we establish a hierarchical video aesthetics evaluation framework that decomposes video aesthetics into three core dimensions, Visual Aesthetics (VA), Visual Fidelity (VF), and Visual Plausibility (VP), with 15 fine-grained criteria. For example, within VA, we define six criteria related to color and lighting, such as color quality and light source, to assess videos.



\begin{figure}[t!]
\centering
\includegraphics[width=1.0\linewidth]{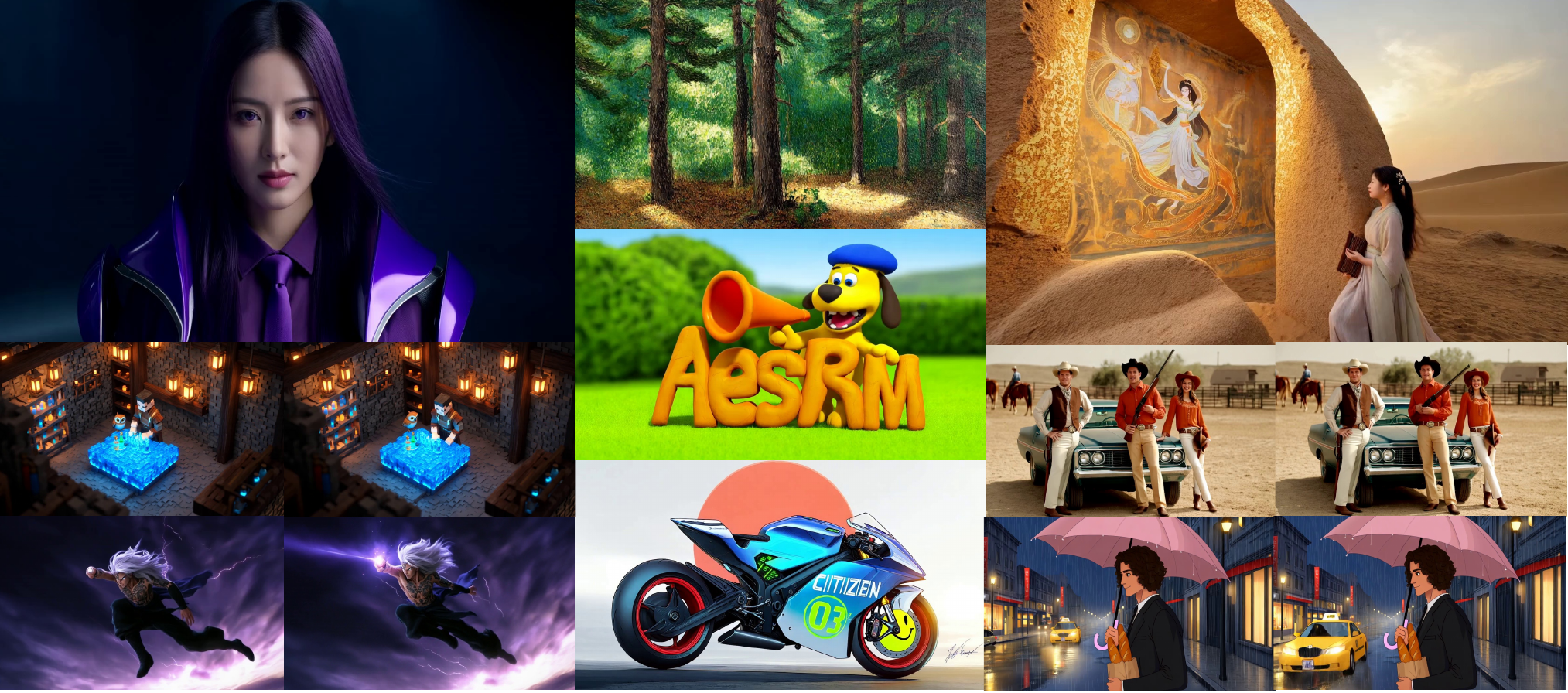}
\caption{Fine-tuning Wan2.2-TI2V-5B \protect\cite{wan2025wan} with AesRM significantly enhances video aesthetics, resulting in more sophisticated lighting, more dynamic shot composition, and richer details. More generated examples are listed in \protect\cref{app:More-Genration-Cases}. }
\label{fig:pipeline}
\end{figure}
Second, based on above hierarchical framework, we collect feedback from multiple aesthetics experts to build a large-scale video aesthetics preference dataset. Specifically, we synthesize approximately 20K high-quality prompts and generate candidate videos using multiple mainstream text-to-video (T2V) models. Aesthetics experts then, guided by the 15 criteria, perform pairwise preference labeling along VA, VF, and VP, selecting the best and worst videos. This pipeline ultimately produces a comprehensive training set of 19K preference pairs and a dedicated video aesthetics benchmark, \textbf{AesVideo-Bench}, comprising roughly 2.5K video pairs with expert fine-grained annotations across VA, VF, and VP.

Then, we develop Video \textbf{Aes}thetic \textbf{R}eward \textbf{M}odels (\textbf{AesRM}), a series of Vision-Language Model (VLM)-based reward models. Instead of adding an auxiliary regression head \cite{ma2025hpsv3,wang2025worldpm}, we cast aesthetic reward modeling as a native next-token prediction task to align with the VLM’s original generative objective, thereby fully leveraging its pre-trained reasoning capabilities \cite{wu2025rewarddance}. Specifically, we design a \textbf{three-stage progressive training strategy} for AesRM:

 \textbf{Stage 1: Atomic Aesthetic Capability Learning}. To enhance general VLMs with domain-specific aesthetic expertise, we first fine-tune AesRM on atomic aesthetics datasets which helps AesRM master core aesthetic concepts, e.g., identifying centered composition, thereby supporting aesthetic evaluation.
    
 \textbf{Stage 2: Cold-Start}. We then train AesRM on constructed aesthetic preference data to instruction-tune reward models for pairwise aesthetic comparison.

 \textbf{Stage 3: Group Relative Policy Optimization (GRPO).} Finally, we apply GRPO \cite{shao2024deepseekmath} to further enhance AesRM's accuracy via verified rewards.

The three-stage pipeline yields AesRM family: \textbf{AesRM-Base} directly outputs pairwise comparison results across VA, VF, and VP, while \textbf{AesRM-CoT} is distilled from Chain-of-Thought (CoT) data  \cite{wei2022chain} generated by Gemini 2.5 Pro \cite{comanici2025gemini25pushingfrontier} and expert priors. The former produces high-efficiency reward signals on core aesthetic dimension for post-training, whereas the latter additionally outputs reasoning based on 15 criteria, improving interpretability of evaluation.  

Notably, we find that low-quality CoT data can amplify hallucinations \cite{augustin2025dash,liu2024survey,liu2025reducing} in AesRM-CoT, e.g., misclassifying a normal video as having severe artifacts. To improve CoT quality, we introduce a self-consistency–based CoT synthesis method that ensembles multiple reasoning trajectories from teacher models using a consistency metric, rather than relying on a single sampled trajectory. This strategy reduces hallucinations in the teacher models and improves the quality of the distilled CoT data. Additionally, to fully exploit the ensembled high-quality CoT data, during GRPO, we incorporate process reward which are formulated based on the similarity between teacher-generated and student-generated CoTs. Extensive evaluations demonstrate the necessity of incorporating ensemble-based CoT and process rewards, enabling our AesRM series to achieve substantial gains across multiple aesthetic benchmarks compared with existing aesthetic models.

Finally, we validate the effectiveness of AesRM in enhancing video aesthetics through two distinct post-training strategies: Flow-RWR \cite{liu2025improving}, an SFT-style reward-weighted regression approach, and Pref-GRPO \cite{wang2025pref}, an RL-based strategy leveraging within-group candidate comparisons. Qualitative results in \cref{fig:pipeline} indicate that AesRM can markedly elevate aesthetic quality by encouraging more refined lighting, more impactful composition, and richer details.

\noindent \textbf{Contribution.} Our contributions are summarized as follows:

{Conceptually}, we hierarchically define video aesthetics by decomposing it into three dimensions and 15 fine-grained criteria (\cref{sec:Hierarchical}). Leveraging this framework, we introduce AesVideo-Bench, comprising approximately 2500 video pairs with expert annotations across VA, VF and VP (\cref{sec:data_collect}).

{Methodology}, we propose a three-stage training to evolve AesRM from basic aesthetic recognition to advanced aesthetic reasoning. This strategy produces AesRM-Base, which provides efficient three-dimensional video evaluations, and AesRM-CoT, which generates additional CoT across 15 criteria (\cref{sec:Three-Stage Progressive Training for AesRM}).

{Insightfully}, we show low-quality distillation data degrades AesRM's accuracy. Consequently, we introduce self-consistency-based CoT synthesis in cold-start and a CoT-based process reward in GRPO to boost AesRM-CoT (\cref{sec:Three-Stage Progressive Training for AesRM}).

{Empirically}, evaluations across multiple aesthetic benchmarks demonstrate AesRM outperforms existing baselines. And integrating AesRM into the alignment of video models significantly enhances visual aesthetics (\cref{sec:exp}).




\section{Related Work}
\label{sec:related work}

\noindent \textbf{Benchmarks for Image and Video Aesthetics.} Recently, Image Aesthetic Assessment (IAA) has advanced rapidly, with many datasets spanning diverse aesthetic dimensions such as clarity, color and shooting view, providing a strong foundation for evaluating image aesthetic models \cite{murray2012ava,kong2016photo,ren2017personalized,fang2020perceptual,hosu2020koniq,achlioptas2021artemis,he2022rethinking,schuhmann2022improved,mavridaki2015comprehensive,cao2025artimuse,cao2025unipercept,liu2025unlocking,bhattacharya2010framework,behrad2025charm,ghosal2022image,ke2023vila,zhu2024attribute,zhang2023blind,liao2025humanaesexpert,zhou2024uniaa}. However, progress in video aesthetics has been limited. Owing to the difficulty of video collection and the expertise required for aesthetic evaluation, existing video aesthetic datasets are mostly restricted to coarse-grained labels \cite{niu2012makes,kuang2019deep,tzelepis2016video} and often adopt a subjective human-preference paradigm \cite{bylinskii2015towards,wu2023exploring,asarkar2019detecting,liu2025improving}, lacking rigorous and systematic evaluation standards \cite{liu2025improving,phatak2019deep,wu2023exploring}. Our work introduces AesVideo-Bench, annotated by aesthetic experts across three core aesthetic dimensions, enabling fine-grained aesthetic evaluation.

\noindent \textbf{Assessments for Image and Video Aesthetics.} A typical approach to improving the quality of images and videos is to leverage external evaluation models to provide reward signals for the generative model, thereby encouraging the generative model to align with the preferences of the aesthetic assessment model \cite{liu2025improving,han2025turning,feng2025video,cao2025vqathinker}. Early image and video aesthetic assessment models often fine-tuned CLIP-style encoders to regress a single aesthetic score \cite{schuhmann2022improved,he2023thinking,qiao2025vadb,gao2024aesmamba}. While lightweight, this paradigm offers limited interpretability and generalization \cite{cao2025artimuse}. More recently, VLM-based evaluators have emerged by fine-tuning VLM backbones on domain-specific datasets \cite{ke2023vila,liao2025humanaesexpert,wu2023q,zhou2024uniaa,zhang2023blind}. However, these methods rely on coarse aesthetic definitions and cannot provide interpretable, fine-grained scores for aesthetic alignment. This motivates AesRM, built on a hierarchical aesthetic rubric, with two variants: AesRM-Base for efficient three-dimensional assessment, and AesRM-CoT for better interpretability via per-criterion reasoning traces.
\section{Video Aesthetic Preference Data Collection}
\label{sec:data}

\subsection{Hierarchical Aesthetic Evaluation Framework}  
\label{sec:Hierarchical}

\begin{wrapfigure}{r}{0.43\textwidth}
  \centering
\includegraphics[width=0.48\textwidth]{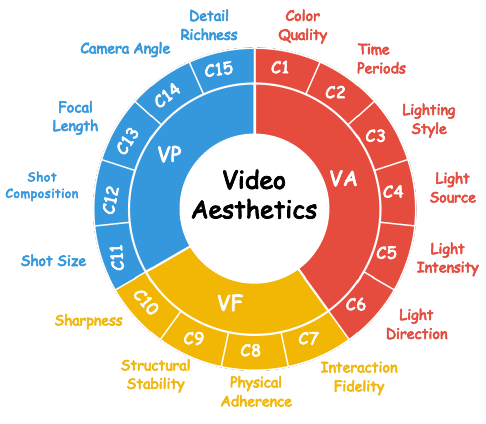}
 \vspace{-20pt}
  \caption{Video aesthetics comprises three core dimensions, \textcolor{lightred}{VA}, \textcolor{lightyellow}{VF} and \textcolor{lightblue}{VP}, and 15 fine-grained criteria}
  \label{fig:cre_aes}
  \vspace{-20pt} 
\end{wrapfigure}

To remedy the lack of systematic definition of video aesthetics, we combine aesthetic expert perspectives with aesthetic concepts from IAA to build the hierarchical evaluation framework in \cref{fig:cre_aes}.  It decomposes video aesthetics into three core aesthetic dimensions and 15 fine-grained criteria as follows:

\noindent \textbf{Visual Aesthetics (VA)} emphasizes video should exhibit harmonious color and cinematic lighting  that faithfully convey the intended atmosphere, e.g., conveying the warm tone through a harmonious orange palette. Focusing on color and light, VA is decomposed into six criteria, including basic light direction recognition (e.g., backlighting) and overall color quality assessment (e.g., oversaturation).

\noindent \textbf{Visual Fidelity (VF)} emphasizes  object interactions in videos should conform to physical commonsense and subjects should remain continuous in shape and texture. Moreover, aesthetic videos should maintain sufficient sharpness to avoid unnatural smearing artifacts. Therefore, VF encompasses four criteria, e.g., structural stability and sharpness.

\noindent \textbf{Visual Plausibility (VP)} focuses on professional cinematic-language, including camera viewpoint and visual richness. Accordingly, VP includes five criteria, like shot composition and focal length.


Collectively, this hierarchical aesthetic framework, comprising three core dimensions and fifteen fine-grained criteria, covers key aesthetic concepts ranging from color and lighting to camera viewpoint, providing a detailed definition of video aesthetics. Detailed descriptions of each criterion can be found in \cref{app:Hierarchical}.

\subsection{Expert-Level Video Aesthetic Preference Data Collection} 
\label{sec:data_collect}

Based on the above aesthetic criteria, we construct an expert-feedback-based aesthetic video preference dataset. Specifically, we first collect a diverse set of raw prompts from the Internet and refine them into approximately 20K high-quality prompts using an in-house prompt rewriter. These prompts typically include both explicit semantic constraints (e.g., \texttt{generate a cat}) and concrete visual-aesthetic instructions (e.g., \texttt{low-angle shot and cool color tones}), providing precise requirements for subsequent aesthetic assessment. We then use four mainstream T2V models to generate about 80K candidate videos from these prompts. Pairing each prompt with its four generated videos yields roughly 20K prompt–video quintuplets.

Then, aesthetic experts perform pairwise comparisons of the candidate videos. They refer to the 15 fine-grained criteria and assign labels across VA, VF, and VP, together with a brief justification. For example, Video A outperforms Video B in VA, while VF and VP are comparable (\texttt{Expert Label}), due to noticeable overexposure in Video B (\texttt{Expert Reason}). In this way, we obtain a video preference dataset in which each prompt is paired with the aesthetically best and worst videos among the candidates. We further partition about 2.5K pairs to establish \textbf{AesVideo-Bench} for video aesthetic evaluation and the remaining pairs are used to train our video aesthetic reward model.

\section{Video Aesthetic Reward Models}
\label{sec:Three-Stage Progressive Training for AesRM}

\begin{figure}[t!]
\centering
  \includegraphics[width=1\textwidth, height=0.09\textheight]{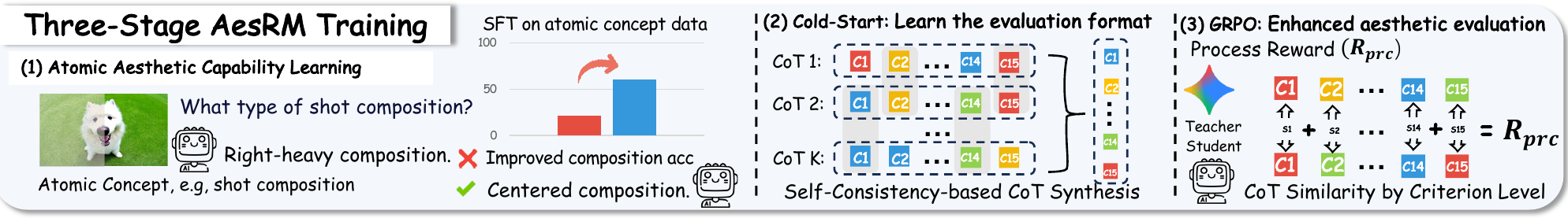}
\caption{Three-Stage Training of AesRM. (1) Atomic Aesthetic Capability Learning improves AesRM's recognition of basic aesthetic concepts, e.g., correctly classifying shot composition as centered via SFT on a curated atomic-aesthetics dataset. (2) Cold-Start stage aligns AesRM with the structured aesthetic-evaluation protocols, notably using self-consistency–based CoT synthesis to improve training data quality. (3) GRPO stage further improves evaluation accuracy of AesRM with additional process rewards.}
\label{fig:training_pipeline}
\end{figure}

We then develop video \textbf{Aes}thetic \textbf{R}eward \textbf{M}odels (AesRM) based on the constructed preference data. Leveraging \texttt{Expert Label} and brief \texttt{Expert Reason}, we train two variants of AesRM: \textbf{AesRM-Base} and \textbf{AesRM-CoT}. The former uses only the \texttt{Expert Label} across VA, VF and VP, enabling fast assessment of video pairs The latter further incorporates the \texttt{Expert Reason}, producing not only the three-dimensional judgment but also an explicit, long CoT reasoning grounded in the 15 fine-grained criteria. We achieve this with a three-stage training pipeline that progressively equips AesRM to recognize fundamental aesthetic concepts and apply them to pairwise video comparisons, as shown in \cref{fig:training_pipeline}.



\subsection{Stage 1: Atomic Aesthetic Capability Learning}
\label{sec:stage1}

Accurate video aesthetic assessment often requires the model to correctly recognize aesthetic attributes in a video, such as brightness and saturation. However, general VLMs typically lack domain-specific aesthetic knowledge, leading to unreliable judgments of basic aesthetic concepts. Taking shot composition as an example, we use an in-house atomic aesthetic image dataset: given an image, the VLM classifies the shot-composition type (see \cref{fig:training_pipeline}) to evaluate the backbone’s ability to recognize basic aesthetic elements. We find even advanced VLMs, like InternVL 3.5~\cite{wang2025internvl3}, achieve only about 20\% accuracy, which severely limits the accuracy of pairwise video aesthetic evaluation. Therefore, we introduce an {atomic aesthetic capability learning} stage, which applies supervised fine-tuning (SFT) to explicitly strengthen the VLM’s recognition of foundational aesthetic concepts.

\begin{figure}[t!]
\centering
  \includegraphics[width=1\textwidth, height=0.15\textheight]{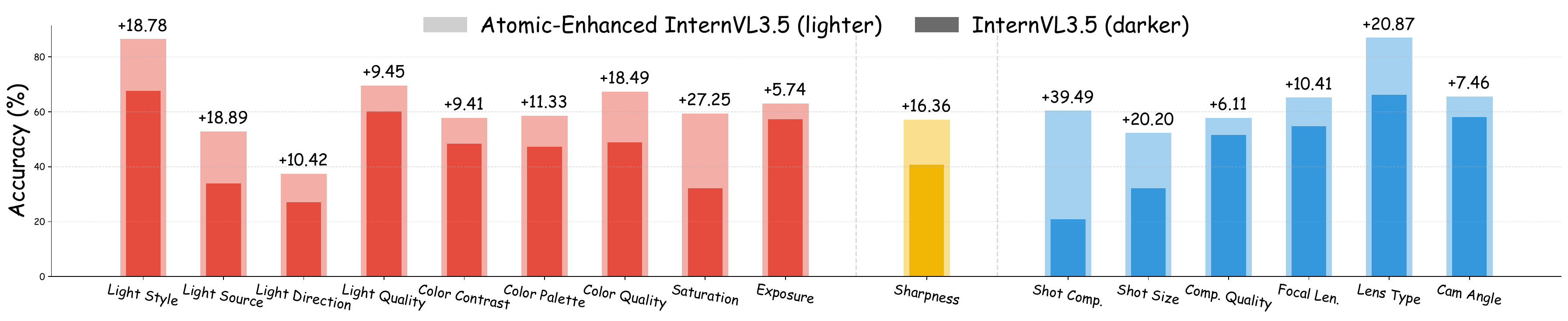}
\caption{Atomic Aesthetic Capability Learning strengthens InternVL 3.5’s ability to recognize fundamental aesthetic concepts with accuracy gains across 16 atomic attributes.}
\label{fig:stage-1-acc}
\end{figure}

Specifically, we distill 16 fundamental aesthetic concepts, e.g., light source and exposure in \cref{fig:stage-1-acc}, from 15 fine-grained criteria as primitives for aesthetic reasoning. Using these concepts, we build an in-house atomic aesthetic dataset by collecting diverse open-source images and obtaining expert annotations for 16 attributes.  Then, we cast atomic concept recognition as an instruction-following task, prompting the model to output discrete labels for each attribute given an image. The VLM is then fine-tuned via supervised next-token prediction using expert annotations as ground truth (see \cref{app:Atomic-Aesthetic-Capability-Learning} for details). As shown in \cref{fig:stage-1-acc}, this stage improves fundamental video aesthetic recognition, raising accuracy on attributes such as saturation and shot composition by 27\%--39\%.
\subsection{Stage 2: Cold-Start for AesRM}
\label{sec:stage2}

This stage aims to equip AesRM with requisite output formats for aesthetic assessment via SFT \cite{guo2025deepseek, wang2025unified, zhang2025thyme}. Specifically, AesRM-Base is supervised to report comparative evaluations across VA, VF, and VP, whereas AesRM-CoT is further tasked with generating explicit reasoning traces grounded in  15 evaluation criteria. To enhance the performance of AesRM-CoT, we introduce an ensemble-based CoT synthesis strategy to improve the quality of the distilled CoT data.

\begin{wrapfigure}{r}{0.48\textwidth}
  \centering
  \vspace{-10pt} 
  \includegraphics[width=0.48\textwidth]{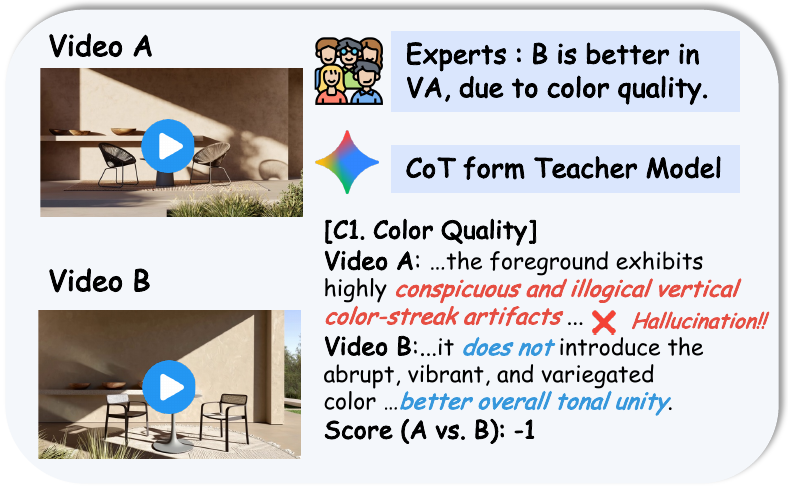}
  \caption{A case where teacher model, i.e., Gemini 2.5 Pro, hallucinated while explaining expert labels, incorrectly identifying video A as having color artifacts.}
  \label{fig:cot_case}
  \vspace{-10pt} 
\end{wrapfigure}
\noindent \textbf{Self-Consistency-based CoT Synthesis.}  To enable AesRM-CoT to reason across the 15 criteria, we require high-quality CoT training data. In practice, we adopt an expert-prior-guided CoT synthesis method that feeds teacher models, i.e., Gemini 2.5 Pro,  with experts' aesthetic priors, \texttt{Expert Label} and  \texttt{Expert Reason}. The teacher is then tasked with expanding these judgments into a step-by-step CoT strictly aligned with the 15 criteria. For instance, if the prior states  Video A outperforms B in VA, while VF and VP are comparable (\texttt{Expert Label}), due to noticeable overexposure in Video B (\texttt{Expert Reason}), the teacher should explain how overexposure-related criteria, e.g., color quality and light intensity, lead to the VA gap, while also justifying why other criteria remain similar, supporting the comparable labels for VF and VP. Compared with from-scratch synthesis without priors, this approach keeps reasoning traces consistent with expert conclusions and improves data utilization efficiency. Detailed prompts for teacher models are in \cref{app:Cold-Start-for-Aesthetic-Evaluation}.

However, we find incorporating expert priors increase the risk of teacher-model hallucination. For example, \cref{fig:cot_case} shows Gemini 2.5 Pro claim input video contains severe color artifacts even when it appears normal, in order to justify expert’s judgment that the video has poor color quality.  To mitigate this issue, we propose self-consistency-based CoT ensembling method shown in \cref{fig:training_pipeline}. 


Formally, we represent the CoT across 15  fine-grained criteria as a set of criterion-level modules \(\{c_i\}_{i=1}^{N=15}\), where each module \(c_i\) consists of a textual rationale \(t_i\) and an evaluation score \(s_i \in \{-1,0,1\}\) (see the output case in \cref{fig:cot_case}). We obtain a reasoning-module pool \(\{c_{j,i}\}_{j=1,i=1}^{M,N}\) by sampling the teacher model \(M\) times for each video pair. And we argue that high-quality CoT data should satisfy the following two requirements: (1) Multi-sample consistency: for the same video pair under the same prompt, rationales from multiple teacher samples should be largely consistent and substantial discrepancies often indicate hallucination \cite{freiberg1983consistency,joo2025black,wang2025joint}; (2) Reasoning–conclusion consistency: criterion-wise judgments in the CoT should support the final dimension-level decision. For example, for video pair \(A\) and \(B\), if \(\sum_{i=1}^{6} s_i < 0\), i.e., the six VA-related criteria indicates $A \succ B$, then the final output should label \(A\) underperforms \(B\) in VA.

Then, we formalize above requirements as the following objective, i.e.,
\begin{equation}
\begin{aligned}
\max_{\{\alpha_{j,i}\}} \quad
& \sum_{j=1}^{M}\sum_{i=1}^{N} \alpha_{j,i}\, f_{j,i} \\
\text{s.t.}\quad
& \sum_{j=1}^{M}\sum_{i=1}^{N} \alpha_{j,i}\, s_{j,i}\in \mathcal{I}_L,\quad
  \sum_{i=1}^{N} \alpha_{j,i}=1.
\end{aligned}
\label{eq:cot}
\end{equation}
where \(\alpha_{i,j}\) denotes the selection probability of the \(j\)-th sampled reasoning for criterion \(i\) and $\mathcal{I}_L$ is determined by \texttt{Expert Label} and defined as
\begin{equation}
\mathcal{I}_L=
\begin{cases}
(0,+\infty), & L:\ A \succ B,\\
(-\infty,0), & L:\ B \succ A,\\
\{0\}, & L:\ A \sim B.
\end{cases}
\end{equation}
And \(f_{j,i}\) is the consistency score of the reasoning for the \(i\)-th criterion produced in the \(j\)-th sample, i.e.,
\begin{equation}
f_{j,i}=\frac{\mathbf{v}_{j,i}\cdot \bar{\mathbf{v}}_{j}}{\|\mathbf{v}_{j,i}\|\,\|\bar{\mathbf{v}}_{j}\|},\qquad
\bar{\mathbf{v}}_{j}=\frac{1}{M}\sum_{k=1}^{M}\mathbf{v}_{j,k},
\end{equation}
where $\mathbf{v}_{j,i}$ is the embedding of text $t_{j,i}$ and extracted via SBERT~\cite{reimers2019sentence}.

\cref{eq:cot} aims to maximize the overall consistency of  synthesized CoT data while satisfying the constraint imposed by the expert label $\mathcal{I}_L$. For such linearly constrained optimization, we treat \(\alpha_{j,i}\in\{0,1\}\) as a hard selection variable for each criterion and perform a discrete search to identify a feasible CoT combination. Detailed optimization procedures are provided in \cref{app:Cold-Start-for-Aesthetic-Evaluation}.

\noindent \textbf{Cold-Start.} Self-consistency-based synthesis enables us to obtain high-quality CoT data infused with expert priors from Gemini 2.5 Pro. We then cold-start AesRM by optimizing a cross-entropy next-token prediction objective on this data, helping it to master the reasoning format for video aesthetic evaluation.
\subsection{Stage 3: GRPO for AesRM}
\label{sec:stage3}
The third stage of AesRM training uses GRPO, a reinforcement learning method that leverages verifiable reward signals to improve AesRM’s accuracy. For AesRM-Base and AesRM-CoT, we design the following reward signals:

\noindent \textbf{Rewards for AesRM-Base} comprise format reward $R_{\mathrm{fmt}}$ and accuracy reward $R_{\mathrm{acc}}$, ensuring AesRM outputs are both well-structured and correct. Specifically,

\textbf{Format reward ($R_{\mathrm{fmt}}$)} encourages AesRM-Base adheres to a strict structural protocol: evaluations must be enclosed within \texttt{<answer>} tags and simultaneously report results across three aesthetic dimensions (VA, VF, and VP). In practice, we use an external judge, Qwen3-8B \cite{yang2025qwen3}, to verify whether the AesRM-Base output follows the required format. We assign \(R_{\text{fmt}}=1\) if the format is satisfied and \(R_{\text{fmt}}=0\) otherwise.

\textbf{Accuracy reward ($R_{\text{acc}}$)} evaluates whether the model's predictions for VA, VF, and VP agree with expert labels. We also use Qwen3-8B to extract the three dimension labels from the output and score each dimension as 1 if it matches the ground truth and 0 otherwise. The accuracy reward is
\begin{equation}
\label{eq:acc_reward}
    R_{\text{acc}}=\mathbb{I}(\text{VA}=\text{VA}_{\mathrm{GT}})+\mathbb{I}(\text{VF}=\text{VF}_{\mathrm{GT}})+\mathbb{I}(\text{VP}=\text{VP}_{\mathrm{GT}}).
\end{equation}
where ${\mathrm{GT}}$ denotes the ground truth, i.e., expert labels.

Finally, the overall verifiable reward for AesRM-Base is formulated as:
\begin{equation}
    R_{\mathrm{AesRM-Base}} = R_{\mathrm{acc}} + \lambda \cdot R_{\mathrm{fmt}},
\end{equation}
where $\lambda > 0$ is a reward weight that balances format and task accuracy.

\noindent \textbf{Rewards for AesRM-CoT} also incorporates format $R_{\mathrm{fmt}}$ and accuracy rewards $R_{\mathrm{acc}}$. Regarding the formatting constraints, AesRM-CoT is required to encapsulate the reasoning within \texttt{<think>} tags, followed by  final evaluation in \texttt{<answer>} tags. Beyond reporting dimensional results, AesRM-CoT should ensure the CoT includes explicit reasoning traces for all 15 fine-grained criteria. More details for $R_{\mathrm{fmt}}$ are provided in \cref{app:GRPO-for-Aesthetic-Evaluation}. Additionally, AesRM-CoT introduces consistency reward $R_{\mathrm{cst}}$ and process reward $R_{\mathrm{prc}}$ as follows:

\textbf{Consistency reward (\(R_{\mathrm{cst}}\))} encourages consistency between the CoT reasoning and the final decision of AesRM-CoT. Specifically, for each dimension, the sum of the fine-grained criterion scores \(\sum_i s_i\) should align with the dimension-level conclusion. For example, for videos \(A\) and \(B\), if \(\sum_{i=1}^{6} s_i > 0\) for the VA dimension, then the final conclusion should be \(A \succ B\) in VA.

\textbf{Process reward ($R_{\mathrm{prc}}$)} is designed to fully exploit the high-quality CoT data synthesized in \cref{sec:stage2}. Specifically, given the same input, for each criterion \(i\), we compute the similarity between the student model’s, i.e., AesRM-CoT, reasoning \(\{t_i, s_i\}\) and the teacher model’s, i.e., Gemini 2.5 Pro, reasoning \(\{t^{\text{teacher}}_i, s^{\text{teacher}}_i\}\). The similarity is a mixture of two metric: (1) the cosine similarity of SBERT embeddings \(R_{\cos}\), capturing overall semantic agreement between the student and teacher reasoning, and (2) the ROUGE-L \cite{lin-2004-rouge} score \(R_{\text{ROUGE-L}}\), measuring lexical overlap and sequence-level alignment. Accordingly, the process reward is the mean mixed similarity as
\begin{equation}
\label{eq:erward_prc}
R_{\text{prc}}
=\frac{1}{N}\sum_{i=1}^{N}
\mathbb{I}\!\left(s_{(i)}^{\text{teacher}}=s_{(i)}^{\text{student}}\right)
\Big(R_{\cos}^{(i)}+R_{\text{ROUGE-L}}^{(i)}\Big).
\end{equation}
where the similarity is considered only when the student and teacher assign the same score , i.e., $s_{(i)}^{\text{teacher}}=s_{(i)}^{\text{student}}$, for $i$-th criterion.

Finally, the overall verifiable reward for AesRM-CoT is formulated as:
\begin{equation}
    R_{\mathrm{AesRM-CoT}} = R_{\mathrm{acc}} + \lambda_{\mathrm{fmt}} \cdot R_{\mathrm{fmt}} + \lambda_{\mathrm{cst}} \cdot R_{\mathrm{cst}} + \lambda_{\text{prc}} \cdot R_{\text{prc}},
\end{equation}
where $\lambda_{\mathrm{fmt}}, \lambda_{\mathrm{cst}}, \lambda_{\mathrm{prc}} > 0$ are reward weights.

\section{Experiments}
\label{sec:exp}

In the experimental section, we primarily focus on: (1) validating the effectiveness of AesRM for video aesthetic assessment (\cref{sec:exp-rm}); (2) evaluating the robustness of AesRM, i.e., whether its aesthetic evaluations are sensitive to factors such as the order of the input video pair (\cref{sec:exp-bias}); and (3) leveraging AesRM for post-training aesthetic alignment in video generation (\cref{sec:exp-video}).

\subsection{Reward Model Performance}
\label{sec:exp-rm}

\noindent \textbf{Training Setup.} We employ InternVL 3.5 \cite{wang2025internvl3} as the backbone for AesRM and follow the three-stage training procedure described in \cref{sec:Three-Stage Progressive Training for AesRM}. For the cold-start stage, we sample 3K instances to construct the ensembled CoT data for AesRM-CoT, and reserve the remaining preference data for GRPO. More details for each stage, including hyperparameters and reward weights, are listed in \cref{app:Three-Stage}.


\begin{table}[tb]
  \caption{Comparison of Aesthetic Reward Models. Binary classification accuracy across multiple aesthetic and human-preference benchmarks are reported. Following \protect\cite{liu2025improving}, w/o tie excludes tie cases. Results show AesRM outperforms existing baselines on AesVideo-Bench and demonstrates strong generalization on OOD data. }
  \label{tab:rm_bench}
  \centering
  \small
  \resizebox{\linewidth}{!}{%
  \begin{tabular}{@{}l c cc cc c cc@{}}
    \toprule
    \textbf{Model (\%)} &
    \textbf{ID Bench} &
    \multicolumn{5}{c}{\textbf{OOD Bench}} &
    \multicolumn{2}{c}{\multirow{2}{*}{\textbf{Avg. $\uparrow$}}} \\ 
    \cmidrule(lr){2-2} \cmidrule(lr){3-7} 
    & {AesVideo $\uparrow$} &
    \multicolumn{2}{c}{VideoGen-Reward $\uparrow$} &
    \multicolumn{2}{c}{GenAI $\uparrow$} &
    {VideoDPO $\uparrow$} & \multicolumn{2}{c}{} \\ 
    \cmidrule(lr){3-4} \cmidrule(lr){5-6} \cmidrule(lr){8-9}
    & & {w/ ties} & {w/o ties} & {w/ ties} & {w/o ties} & & {w/ ties} & {w/o ties} \\
    \midrule

    \multicolumn{9}{l}{\textit{Image Aesthetics}\vspace{0.02in}} \\
    LAP \cite{schuhmann2022improved}
      & 56.17 & \underline{34.26} & \underline{65.14} & 39.07 & 58.30  & \underline{58.28} & 46.95 & 59.47 \\
    ArtiMuse \cite{cao2025artimuse}
      & 52.19 & 30.76 & 57.44 & 41.83 & 62.31 & 50.30 & 43.77 & 55.56 \\
    \addlinespace[1pt]

    \multicolumn{9}{l}{\textit{Video Aesthetics}\vspace{0.02in}} \\
    VADB \cite{qiao2025vadb}
      & 52.56 & 28.69 & 54.56 & 37.26 & 55.57 & 52.79 & 42.83 & 53.87 \\
    VideoAlign-VQ \cite{liu2025improving}
      & 52.89 & \textbf{39.81} & \textbf{75.30} & 40.84 & 61.19 & 51.03 & 46.14 & 60.10 \\
    \grayrow
    AesRM-Base
      & \underline{68.86} & 29.00 & 51.55 & \underline{45.21} & \underline{67.44} & \textbf{58.46} & \underline{50.38} & \underline{61.58} \\
    \grayrow
    AesRM-CoT
      & \textbf{69.41} & 31.87 & 59.65 & \textbf{46.29} & \textbf{68.89} & 57.56 & \textbf{51.29} & \textbf{63.88} \\
    \bottomrule
  \end{tabular}%
  }
\end{table}

\noindent \textbf{Evaluation Benchmarks.} We categorize evaluation benchmarks into: (1) In-Distribution (ID), where we use AesVideo-Bench, consisting of pairwise video assessments annotated by aesthetic experts along three core dimension and (2) Out-of-Distribution (OOD), including VideoGen-RewardBench \cite{liu2025improving,zeng2024dawn}, GenAI-Bench \cite{jiang2024genai}, and VideoDPO \cite{liu2025videodpo}. Unlike AesVideo-Bench, these OOD benchmarks rely on general human preferences rather than fine-grained aesthetic definitions. For VideoGen-RewardBench, we adopt the Visual Quality (VQ) dimension as ground truth due to its alignment with video aesthetics. In both ID and OOD settings, reward models predict the better video in each pair, and the binary accuracy are reported as metric. More introductions of benchmarks are in \cref{app:Evaluation}.

\noindent \textbf{Baselines.} Given the limited video aesthetic reward models, we also include IAA models as complementary baselines. And baselines include: (1) image aesthetic models, including the LAION Aesthetic Predictor (LAP) \cite{schuhmann2022improved}, trained on human ratings using frozen CLIP \cite{radford2021learning} embeddings, and ArtiMuse \cite{cao2025artimuse}, a VLM-based model that outputs continuous aesthetic scores for images and (2) video aesthetic baselines, including VADB-Net \cite{qiao2025vadb}, a multimodal CLIP model fine-tuned on the self-constructed aesthetics dataset \cite{qiao2025vadb}, and VideoAlign \cite{liu2025improving}, a multi-dimensional reward model that evaluates visual quality, motion quality, and text alignment. For VideoAlign, we use its aesthetic-related visual quality score as prediction. More details on the baselines are provided in \cref{app:Baselines}.

\noindent \textbf{Main Results.} \cref{tab:rm_bench} evaluates AesRM, i.e., AesRM-Base and AesRM-CoT, against other aesthetic baselines across multiple benchmarks. AesRM performs strongly on ID benchmarks, improving accuracy by at least 12\% over the best baseline and reaching nearly 70\%. This result indicates that AesRM is strongly aligned with expert aesthetic preferences. AesRM also generalizes well to OOD benchmarks, even achieving the best results in  GenAI-Bench and VideoDPO.



\subsection{Reward Model Position Bias}
\label{sec:exp-bias}

In large language models (LLMs),  position bias  refers to the tendency to favor solutions based on their position within the prompt \cite{wang2024eliminating,bito2025evaluating,shi2025judging}. For AesRM, this corresponds to whether its judgments are sensitive to the order of the input video pair. To assess the bias, we perform bidirectional inference for each video pair and examine the consistency of its preferences. Specifically, given a video pair $(V^{(1)}, V^{(2)})$ with prompt $\mathcal{P}$, we define the position-bias score as:
\begin{align*}
\label{eq:position_bias}
\resizebox{\linewidth}{!}{$
\mathrm{PB}
=\mathbb{E}_{(V^{(1)},V^{(2)},\mathcal{P})}\!
\left[
\mathbb{I}\!\left(
\mathrm{sgn}\!\big(\mathrm{AesRM}(V^{(1)},V^{(2)},\mathcal{P})\big)=
\mathrm{sgn}\!\big(\mathrm{AesRM}(V^{(2)},V^{(1)},\mathcal{P})\big)
\right)
\right]
$}
\end{align*}
\begin{wraptable}{r}{0.43\linewidth}
  \captionsetup{font=small,skip=4pt}
  \caption{AesRM shows robust evaluation with low positional bias.}
  \label{tab:pbscore}
  \centering
  \resizebox{\linewidth}{!}{%
    \setlength{\tabcolsep}{4pt}%
    \renewcommand{\arraystretch}{1.05}%
    \begin{tabular}{@{}lcc@{}}
      \toprule
      \textbf{Model (\%)} & \textbf{AesVideo $\downarrow$} & \textbf{GenAI $\downarrow$} \\
      \midrule
      UniRM \cite{wang2025unified-1} &29.89 &10.44\\
      AesRM-Base & 16.42 & 10.54 \\
      AesRM-CoT  & 21.12 & 10.76 \\
      \bottomrule
    \end{tabular}%
  }
  \vspace{-10pt}
\end{wraptable}

where $\mathrm{sgn}(\cdot)$ denotes the preference direction, e.g., positive sign indicates $A \succ B$. 
$\mathrm{PB}$ measures the fraction of pairs whose preference flips after swapping the input order where an order-invariant reward model should yield $\mathrm{PB}$ close to $0$. 
\cref{tab:pbscore} reports PB scores of AesRM-Base and AesRM-CoT on AesVideo-Bench and GenAI-Bench. For comparison, we also report position bias scores for general pairwise reward models, UniRM \cite{wang2025unified-1}. Results show AesRM consistently achieves PB below 20\%, even dropping to around 10\% on GenAI-Bench, indicating  robust video-pair aesthetic evaluation.

\subsection{Video Aesthetic Alignment}
\label{sec:exp-video}

\noindent \textbf{Training Setup.} We further leverage AesRM for post-training Wan2.2-TI2V-5B \cite{wan2025wan} based on AesVideo-Bench. We mainly focus on Flow-RWR \cite{liu2025improving,peters2007reinforcement}, a weighted SFT method for flow-based models where sample weights are proportional to reward scores and report Pref-GRPO \cite{wang2025pref}, which performs pairwise comparisons within each sample group, in \cref{app:Pref-GRPO}. Details including Flow-RwR, Pref-GRPO, and hyperparameters, are provided in \cref{app:exps}.


\noindent \textbf{Baselines and Evaluation.} Following \cref{sec:exp-rm}, we use LAP, ArtiMuse, and VideoAlign-VQ as baseline reward models for post-training Wan2.2. And we evaluate aesthetic quality of generated videos from two perspectives: (1) Model-based evaluation, using multiple reward models, including the aesthetic reward models in \cref{sec:exp-rm} and general reward model UniRM \cite{wang2025unified-1}, to quantify aesthetic improvements and (2) Expert-based evaluation conducting user study where multiple aesthetic experts perform video comparisons. 

For LAP, ArtiMuse, VideoAlign-VQ, and UniRM, which output scalar scores for individual videos, we utilize the difference in mean scores before and after fine-tuning as quantitative metric. For AesRM-Base and AesR-CoT, which facilitate pairwise video comparisons, we adopt the Good–Same–Bad (GSB) score \cite{wu2025rewarddance}:
\begin{equation}
\label{eq:GSB}
\mathrm{GSB}=\frac{G-B}{G+S+B},
\end{equation} 
where $G$, $S$, and $B$ denote the counts of pairs where the fine-tuned model is judged as better, same, or worse than the pre-trained baseline, respectively.



\begin{table}[tb]
  \captionsetup{font=small,skip=6pt}
  \caption{Video aesthetic evaluation of Wan2.2 fine-tuned with various aesthetic reward models. AesRM yields significantly greater aesthetic improvements than baselines.}
  \label{tab:wan_rwr}
  \centering
  \resizebox{\linewidth}{!}{%
  \begin{tabular}{@{}lccccccc@{}}
    \toprule
    \textbf{Model} & \textbf{LAP $\uparrow$} & \textbf{ArtiMuse $\uparrow$} & \textbf{VideoAlign-VQ $\uparrow$} &
    \textbf{AesRM-Base $\uparrow$} & \textbf{AesRM-CoT $\uparrow$} & \textbf{UniRM \cite{wang2025unified-1} $\uparrow$} & \textbf{Avg. Rank $\downarrow$} \\
    \midrule
    Wan2.2-TI2V-5B            & 5.505 & 56.161 & -0.592 & --     & --     & 2.817 & -- \\
    + \textit{SFT}            & 5.613 & 56.917 & -0.493 & \underline{3.659}  & 18.704 & 2.892 & 3.17 \\
    + \textit{LAP} \cite{schuhmann2022improved}           & 5.615 & 56.814 & -0.497 & -1.518 & 17.141 & 2.892 & 3.83 \\
    + \textit{ArtiMuse}   \cite{cao2025artimuse}    & 5.614 & \textbf{57.074} & -0.512 & -4.944 & 12.656 & 2.885 & 4.50 \\
    + \textit{VideoAlign-VQ} \cite{liu2025improving}  & 5.608 & \underline{57.070} & -0.512 & -0.584 & 15.346 & 2.882 & 4.67 \\
    \grayrow
    + \textit{AesRM-Base}     & \underline{5.621} & 56.614 & \underline{-0.485} & \textbf{5.956}  & \underline{31.721} & \underline{2.910} & \underline{2.33} \\
    \grayrow
    + \textit{AesRM-CoT}      & \textbf{5.623} & 56.494 & \textbf{-0.471} & 1.362  & \textbf{33.854} & \textbf{2.911} & \textbf{2.17} \\
    \bottomrule
  \end{tabular}%
  }
\end{table}


\begin{figure}[t!]
\centering
  \includegraphics[width=1\linewidth]{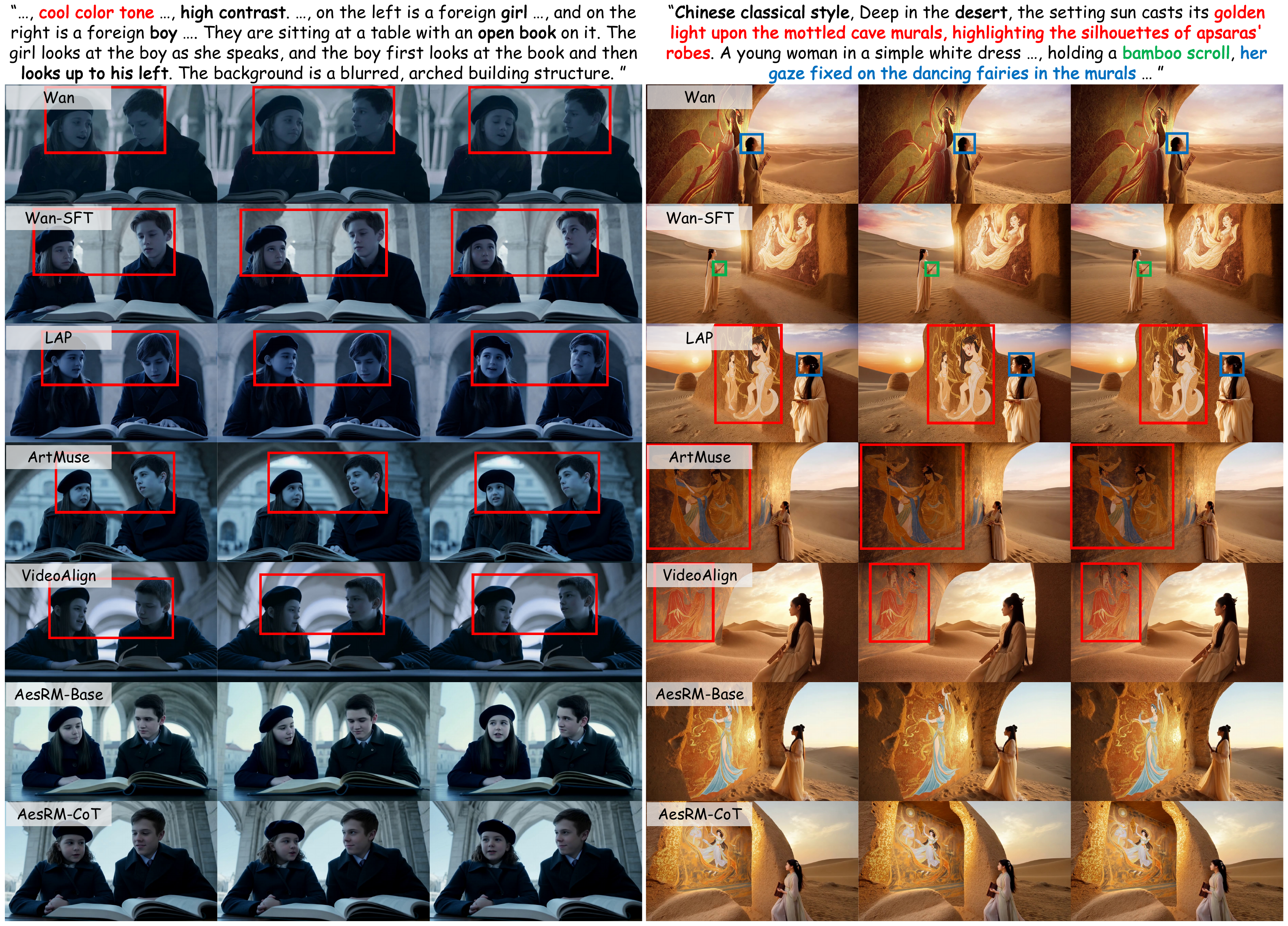}
\caption{Visualization of Wan2.2 generations under different aesthetic reward models. Fine-tuning with AesRM-Base or AesRM-CoT yields better aesthetics. Left: AesRM improves cool-tone color quality while keeping the face natural, avoiding the baseline’s over-darkening. Right: AesRM better matches the prompt’s golden lighting and produces richer details, e.g., the bamboo scroll.}
\label{fig:compare}
\end{figure}

\noindent \textbf{Main Results.} \cref{tab:wan_rwr} shows the aesthetic-alignment gains achieved on Wan2.2 using different aesthetic reward models. We find that AesRM-CoT and AesRM-Base improve aesthetics more effectively than other baselines, achieving the best average ranks of 2.17 and 2.33, respectively. 
\begin{wrapfigure}{r}{0.43\linewidth}
  \centering
  \includegraphics[width=\linewidth]{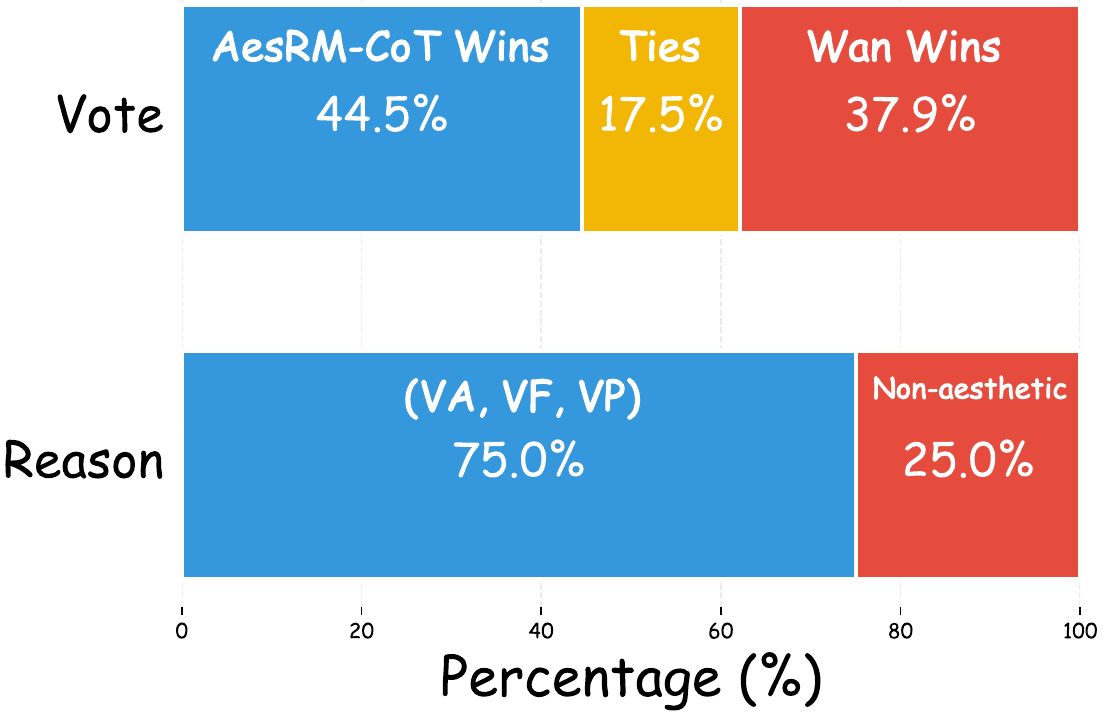}
  \captionsetup{font=small}
  \caption{Experts evaluate videos from fine-tuned Wan2.2 with AesRM-CoT: 45\% of samples show improvements, mainly (75\%) due to better visual aesthetics and composition.}
  \label{fig:user_study}
   \vspace{-10pt}
\end{wrapfigure}
\cref{fig:compare} further provides a qualitative video comparison, where Wan2.2 fine-tuned with AesRM produces videos with better tone, lighting, and detail, without the obvious underexposure or overexposure observed with other reward models. Results of Pref-GRPO are in \cref{app:Pref-GRPO}, which consistently support the effectiveness of AesRM for aesthetic alignment.

\noindent \textbf{User Study.} To assess whether AesRM improves video aesthetics, we conduct a user study on 1.2k videos randomly sampled from AesVideo-Bench. Aesthetic experts perform pairwise comparisons of Wan2.2 videos before and after Flow-RwR and provide brief reasons for their judgments e.g., improved visual aesthetics, better composition, or smoother motion. \cref{fig:user_study} shows Wan fine-tuned with AesRM-CoT yields noticeable improvements on about 45\% of generations, and improvements are primarily (75\%) driven by aesthetic factors, such as better visual aesthetics and composition.


\section{Discussion}
\label{sec:dis}


This section addresses two questions: (1) Is atomic aesthetic capability learning stage in \cref{sec:Three-Stage Progressive Training for AesRM} necessary? (2) For AesRM-CoT, do self-consistency CoT synthesis and process reward in GRPO improve performance?

\begin{wraptable}{r}{0.43\linewidth}
\vspace{-10pt}
\centering
\caption{Improving CoT quality and introducing process rewards boost AesRM-CoT’s performance.}
\label{tab:aba-cot}
\resizebox{\linewidth}{!}{
\begin{tabular}{@{}lcc@{}}
  \toprule
  {Model (\%)} & {Binary} & {Avg.}  \\
  \midrule
  InternVL3.5 & 59.67 & 51.95  \\
  + \textit{Cold-Start and GRPO} & 65.51 & 64.87 \\
  \enspace + \textit{Ensemble CoT} & 66.61 & 64.97  \\ 
  \enspace \enspace + \textit{Process Reward} & \textbf{66.99} & \textbf{65.39} \\
  \bottomrule
  \vspace{-20pt}
\end{tabular}%
} 
\vspace{-0.8\baselineskip}
\end{wraptable}
\noindent \textbf{Atomic Aesthetic Capability Learning.} \cref{tab:aba-aesrm_base_s1} and \cref{tab:aba-aesrm_cot_s1} report ablation of stage 1 on AesVideo-Bench. For both AesRM-Base and AesRM-CoT, atomic capability learning consistently improves evaluation accuracy by 2\%. In addition, \cref{fig:aba-pbscore} shows that enabling Stage 1 (S1) reduces positional bias by 6\%, indicating more robust evaluation. Overall, learning fundamental aesthetic concepts before aesthetic reasoning benefits both accuracy and evaluation robustness.


\begin{table*}[t]
\centering
\setlength{\tabcolsep}{3pt} 
\renewcommand{\arraystretch}{1.05}
\begin{minipage}[t]{0.32\linewidth}
  \centering
  \small 
  \caption{Stage 1 improves accuracy of AesRM-Base where Avg. denotes the mean accuracy over VA, VF, and VP.}
  \label{tab:aba-aesrm_base_s1}
  \resizebox{\linewidth}{!}{%
    \begin{tabular}{@{}lcc@{}}
      \toprule
      \textbf{Model (\%)} & \textbf{Binary} & \textbf{Avg.} \\
      \midrule
      InternVL3.5               & 59.67 & 51.95 \\
      +\textit{Cold-Start and GRPO}   & 66.61 & 64.42 \\
      \enspace+\textit{Atomic Learning} & \textbf{68.86} & \textbf{65.38} \\
      \bottomrule
    \end{tabular}%
  }
\end{minipage}\hfill
\begin{minipage}[t]{0.32\linewidth}
  \centering
  \small
  \caption{Stage 1 improves evaluation accuracy of AesRM-CoT where Avg. denotes the mean accuracy over VA, VF, and VP.}
  \label{tab:aba-aesrm_cot_s1}
  \resizebox{\linewidth}{!}{%
    \begin{tabular}{@{}lcc@{}}
      \toprule
      \textbf{Model (\%)} & \textbf{Binary} & \textbf{Avg. } \\
      \midrule
      InternVL3.5               & 59.67 & 51.95 \\
      +\textit{Cold-Start and GRPO}    & 66.99 & 65.39 \\
    \enspace+\textit{Atomic Learning} & \textbf{69.41} & \textbf{64.88} \\
      \bottomrule
    \end{tabular}%
  }
\end{minipage}\hfill
\begin{minipage}[t]{0.32\linewidth}
  \centering
  \captionsetup{type=figure,font=small,skip=2pt}
  \includegraphics[width=\linewidth,height=3.0cm,keepaspectratio]{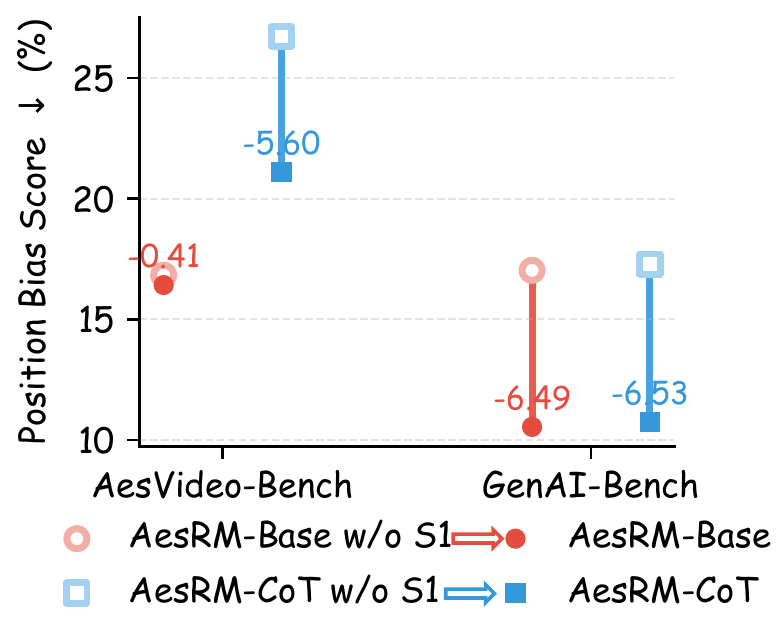}
  \vspace{-2pt} 
  \captionof{figure}{Stage 1 (S1) improves AesRM’s robustness.}
  \label{fig:aba-pbscore}
\end{minipage}
\end{table*}

\noindent \textbf{Self-Consistency-based CoT and Process Rewards.} \cref{tab:aba-cot} shows ensemble CoT and reward processing improve AesRM-CoT. Moreover, comparing \cref{tab:aba-cot} with \cref{tab:aba-aesrm_base_s1}, AesRM-CoT trained with only cold start and GRPO (without process rewards) underperforms AesRM-Base (65.51 vs. 66.61 in binary accuracy), while ensemble CoT mitigates this gap. This indicates for long-CoT tasks like aesthetic assessment, AesRM-CoT is highly sensitive to CoT quality: with low-quality CoT, the non-CoT AesRM-Base baseline may perform better.



\section{Conclusion and Limitation}
\label{sec:conclusion}
Our work systematically advance video aesthetics by hierarchically defining aesthetic with three core dimensions and 15 fine-grained criteria, and building corresponding benchmark, AesVideo-Bench. Leveraging these resources, we develop AesRM-Base and AesRM-CoT via three-stage training with self-consistent CoT synthesis and process reward, to support video aesthetic alignment. This work has the following limitations. Experiments use only 20K samples and a single backbone InternVL 3.5, without studying scaling over data and backbones. In addition, while with expert priors, most CoT is generated by a teacher model. Fully expert-written CoT across all 15 criteria may further improve AesRM.

\section*{Acknowledgement}
We would like to express our gratitude to the aesthetic experts, Danying Wang (Tianjin Academy of Fine Arts), Wenyu Yang (Tianjin University of Technology), Han Wang (Southwestern University of Finance and Economics), Liang Yan (Hubei Institute of Fine Arts), Ceres Su (Central Academy of Fine Arts), and Rita Li (Xi’an Academy of Fine Arts), for providing professional video aesthetic definitions and feedbacks for this project.

\bibliographystyle{iclr2026_conference}
\bibliography{iclr2026_conference}

\clearpage
\appendix
\clearpage

\providecommand{\authcount}[1]{}



\begin{center}
  \Large\textbf{Appendix of AesRM: \\ Improving Video Aesthetics with Expert-Level Feedback}
\end{center}


\begingroup
\hypersetup{hidelinks}

\noindent\textbf{Contents}\par\smallskip
{\small
\noindent A\quad \hyperref[app:Hierarchical]{Hierarchical Aesthetic Evaluation Framework}\dotfill\pageref{app:Hierarchical}\par
\noindent\hspace{2em}A.1\quad \hyperref[app:va]{Visual Aesthetics (VA)}\dotfill\pageref{app:va}\par
\noindent\hspace{2em}A.2\quad \hyperref[app:vf]{Visual Fidelity (VF)}\dotfill\pageref{app:vf}\par
\noindent\hspace{2em}A.3\quad \hyperref[app:vp]{Visual Plausibility (VP)}\dotfill\pageref{app:vp}\par

\noindent B\quad \hyperref[app:Expert-Leve]{Expert-Level Video Aesthetic Preference Data Collection}\dotfill\pageref{app:Expert-Leve}\par

\noindent C\quad \hyperref[app:Three-Stage]{Three-Stage Training of AesRM}\dotfill\pageref{app:Three-Stage}\par
\noindent\hspace{2em}C.1\quad \hyperref[app:Atomic-Aesthetic-Capability-Learning]{Stage 1: Atomic Aesthetic Capability Learning}\dotfill\pageref{app:Atomic-Aesthetic-Capability-Learning}\par
\noindent\hspace{2em}C.2\quad \hyperref[app:Cold-Start-for-Aesthetic-Evaluation]{Stage 2: Cold-Start for Aesthetic Evaluation}\dotfill\pageref{app:Cold-Start-for-Aesthetic-Evaluation}\par
\noindent\hspace{2em}C.3\quad \hyperref[app:GRPO-for-Aesthetic-Evaluation]{Stage 3: GRPO for Aesthetic Evaluation}\dotfill\pageref{app:GRPO-for-Aesthetic-Evaluation}\par

\noindent D\quad \hyperref[app:exps]{Experiments}\dotfill\pageref{app:exps}\par
\noindent\hspace{2em}D.1\quad \hyperref[app:Evaluation]{Evaluation Benchmarks and Metrics}\dotfill\pageref{app:Evaluation}\par
\noindent\hspace{2em}D.2\quad \hyperref[app:Baselines]{Aesthetic Baselines}\dotfill\pageref{app:Baselines}\par
\noindent\hspace{2em}D.3\quad \hyperref[app:RWR]{Flow-RWR}\dotfill\pageref{app:RWR}\par
\noindent\hspace{2em}D.4\quad \hyperref[app:Pref-GRPO]{Pref-GRPO}\dotfill\pageref{app:Pref-GRPO}\par
\noindent\hspace{2em}D.5\quad \hyperref[app:More-Results]{More Quantitative Results}\dotfill\pageref{app:More-Results}\par
\noindent\hspace{2em}D.6\quad \hyperref[app:More-Genration-Cases]{More Qualitative Cases}\dotfill\pageref{app:More-Genration-Cases}\par
}
\endgroup

\noindent Our supplementary material is organized into four parts. 

\begin{enumerate}
    \item \cref{app:Hierarchical} provides a detailed definition of video aesthetics, including three core dimensions and fifteen criteria, along with detailed explanations of each criterion.

    \item \cref{app:Expert-Leve} provides further details on the construction of AesVideo-Bench.
    
      \item \cref{app:Three-Stage} describes the three-stage training pipeline for AesRM, including the system prompts for each stage, full training hyperparameters, and detailed explanations of our key techniques, Self-Consistency-based CoT Synthesis and process reward design. 
      
      \item \cref{app:exps} presents the experimental setup, including evaluated benchmarks, metrics, baselines, and the used two post-training strategies: flow-RWR and pref-GRPO. This section also reports additional quantitative results, including more accuracy metrics for AesRM on AesVideo-Bench and the post-training performance of pref-GRPO, as well as additional qualitative video generations demonstrating improved video aesthetics after applying AesRM.

\end{enumerate}


\newpage

\section{Hierarchical Aesthetic Evaluation Framework}
\phantomsection
\label{app:Hierarchical}

In this section, we discuss in detail the definition of video aesthetics, i.e., the three core dimensions and the 15 aesthetic criteria. Specifically, we provide detailed explanations of above concepts in each dimension as follows:

\subsection{Visual Aesthetics (VA)}
\label{app:va}

This dimension evaluates the artistry and emotional impact of visual content, mainly from two aspects: (1) fundamental image quality and color, and (2) the use of lighting and shadow. An aesthetically pleasing image should, while adhering to physical laws and artistic principles, achieve rich light--shadow layering, harmonious color, and accurately convey the atmosphere and emotion required by the prompt. Unless the prompt explicitly calls for a stylized look (e.g., high-exposure style, dark silhouette), basic image-quality issues, such as overexposure, underexposure  or excessive saturation, at any time of day or in any lighting environment should be considered non-compliant with aesthetic standards. 

\textbf{C1. Color Quality.} This is the foundation of aesthetic evaluation.

\begin{enumerate}
    \item \textbf{Brightness and Contrast}: Brightness refers to the overall lightness/darkness of the image. Contrast refers to the range of difference between the brightest areas (highlights) and the darkest areas (shadows).

    The \textbf{Key checks} include (1) Is the brightness appropriate? Are there issues such as overall too dark, somewhat dark, too bright, or overexposed? For example, in a daytime outdoor scene, the subject's face is completely black due to underexposure, while the sky is blown out into flat white due to overexposure; (2) Is the contrast appropriate? High contrast can enhance depth and visual impact but may lose detail; low contrast can look softer but may appear flat. Does the contrast match the atmosphere required by the prompt (e.g., dramatic vs.\ soft scenes)? (3) How is the dynamic range; is detail lost? Good performance preserves rich tonal layers and details in both highlights and shadows even in high-contrast scenes?

    \item \textbf{Saturation:}  The purity or vividness of color. High saturation appears vivid; low saturation appears muted.

    The \textbf{Key checks} include (1) Is saturation imbalanced? Is the image under-saturated (gray, lifeless) or over-saturated (overly intense, losing detail and layering, looking cheap or unreal)? (2) Does saturation fit the scene? For example, cyberpunk often needs highly saturated neon colors, while a rainy-day scene should be low-saturation.

    \item \textbf{Color harmony:} This focuses on relationships among multiple colors in the image and their influence on overall mood and visual focus. Good color pairing guides emotion, highlights the subject, and creates a harmonious or impactful visual experience.

    The \textbf{Key checks} include (1) Are color combinations harmonious and unified? Are there abrupt or muddy/dirty colors that hurt aesthetics? Does the overall tone (warm/cool, high/low saturation) successfully create the atmosphere required by the prompt (e.g., cozy, chilly, cyberpunk)? (2) Is there a clear hierarchy (primary, secondary, accent colors)? Does the image effectively emphasize the subject via complementary contrast or high-saturation accents? (3) If the prompt specifies a color style (e.g., Morandi palette, autumn warm tones), is it implemented accurately? Is there improper color cast? Unless intentionally stylized, avoid unnatural color shifts, e.g., under normal daylight, white objects should not skew green or purple.
\end{enumerate}

\textbf{C2. Time Periods.} 
\begin{enumerate}
    \item \textbf{Daytime}: Overall bright image, natural/true color, moderate saturation. In sunny daylight there should be clear shadows and higher contrast; in overcast daylight the light should be soft with lower contrast. Performance is poor if the image is inexplicably dark, highlights overflow due to overexposure, or colors are severely distorted.

     \item \textbf{Night} includes (1) Pure outdoor: Large areas of deep dark tones; no sun as key light. Colors are mainly determined by artificial light sources (streetlights, headlights, billboards), with obvious influence from those lights. (2) Indoor with outdoor elements: Indoor scene but with visible outside through windows/doors: outdoor parts must match pure outdoor night characteristics. If the indoor subject is lit by a small concentrated source, expect strong light--dark contrast; if lit by a large-area source (e.g., ceiling light), the environment is brighter and contrast is relatively softer. (3) Lighting should be entirely dominated by indoor artificial sources.

     \item \textbf{Sunset} requires low-angle light direction; warm hues dominated by orange/red/yellow. The sky often shows strong warm vs.\ blue/purple contrast. Ground objects are tinted warm and cast long shadows. Poor if lighting is too bright (like noon), too dark (like late night), or lacks the signature warm-toned sky.

      \item \textbf{Dusk} is between sunset and night. Warm sky tones (golden/orange-yellow) fade while cool tones (blue/purple) dominate. Overall darker than sunset; sky transitions from bright to dark. Poor if still as bright as daytime or lacks dusk color characteristics.

       \item \textbf{Dawn} is the  transition from night to day; sun not yet risen. Light is extremely soft and diffused with almost no shadows. Typically cool tones (blue/magenta), low overall brightness, quiet and mysterious mood. Poor if lighting is too strong with direct beams, or if the mood is indistinguishable from overcast day or night.

       \item \textbf{Sunrise} means sun has just reached or is about to cross the horizon. Very low-position light; slightly bright yellow/orange; begins to produce soft long shadows. Overall brightness rises; shadows may still retain cool dawn tones, creating warm--cool contrast. Poor if too dark, overexposed, or tonally inconsistent.
\end{enumerate}

\textbf{C3. Lighting Style.} 
\begin{enumerate}
    \item \textbf{Practical light}: visible in-frame light sources that viewers can clearly see are emitting light; not hidden production lights), e.g., desk lamps, candlelight, flashlights, candles, TVs.

    The \textbf{Key checks} include does the illumination feel realistic and integrated with the environment, successfully creating the prompt-required mood? Does it model depth? Under good scene lighting, objects show natural light and shadow transitions and volume; poor scene lighting makes objects look flat and lacking spatial layers.

    \item \textbf{Scene lighting}: artificial lighting arranged on set to create overall mood and shape; the source is not visible, but the effect looks plausible as if from a window/skylight/hidden practical.

    The \textbf{Key checks} include does the illumination feel realistic and integrated with the environment, successfully creating the prompt-required mood? Does it model depth? Under good scene lighting, objects show natural light--shadow transitions and volume; poor scene lighting makes objects look flat and lacking spatial layers.
\end{enumerate}

\textbf{C4. Light Source.} The light source is first determined by whether the prompt calls for natural or artificial light; then we check whether the video meets the aesthetic standard under that requirement.

\begin{enumerate}
    \item \textbf{Natural light} include

    (1) Sunny daylight: Usually requires direct sunlight, abundant light, clear visibility, and a slightly warmer mood than neutral daylight. Poor sunny light often shows overexposure or unreasonable overly dark areas.
    
    (2) Overcast daylight: Soft light, low contrast, soft or even absent shadows; weaker edge definition; medium brightness; lower saturation; neutral-to-cool tone; weaker contrast than sunny light, with more visible shadow detail than sunny light.

    (3) Moonlight: Much weaker than daylight; overall darker; higher color temperature with clearly cool tone (blue/cyan). Shadows can exist but are very low-contrast and only faintly visible, with even lower contrast than overcast light.

    (4) Snow light: Overall saturation decreases; light is relatively soft. If people appear, skin tones become cooler; reflected light from snow brightens areas under the nose and jaw, so shadows at the bottom of facial features are less pronounced than under typical natural light.

    \item \textbf{Artificial light} include

    (1) Firelight: Alternating bright/dim, flickering illumination; shadows constantly change and jitter; warm tone (orange-yellow). Stronger firelight makes the image warmer. When near a face with limited coverage, facial contrast is strong.

    (2) Incandescent light: Typical warm yellow; stable; stronger than firelight; sharp shadow edges (hard light); creates a cozy or retro mood.

    (3) Fluorescent light: Often from strip fixtures; the emitter core may appear dead white with higher purity around it. Can cause color distortion and greenish casts; low contrast; relatively flat illumination regardless of range; typically cool. Lower color purity than neon; more like weak colored-but-whitish light.

    (4) Neon light: Highly saturated, high-purity colored light, often as text or graphics. Small-area intense light with strong color spill; clear shadows; often conveys strong emotion, urban fashion, tech, or a sense of vice.

    (5) LED light: Controllable artificial source with strong directionality; high fidelity color reproduction, no flicker, almost no color cast; adjustable warm/cool; a versatile simulated source. Good LED lighting should be concentrated (areas exist that are not color-shifted by it) and have high saturation.

    (6) Mixed light: Combination of different light-source effects, often warm + cool mix; may also include natural + artificial simultaneously; inconsistent directions; multiple shadows (directions may differ). Good mixed light shows rich color with balanced warm/cool.

    (7) High-intensity light: High-energy, strong light that significantly enhances contrast and detail; used for strong visual impact. Good high-intensity light is strong and relatively concentrated, producing large bright--dark contrast.
    
\textbf{Note:} Determine whether the prompt requires natural or artificial light and which subtype, then judge accordingly. Regardless of type, images that are too bright or too dark should be considered not meeting aesthetic standards.
    
\end{enumerate}

\textbf{C5. Light intensity.} 

\begin{enumerate}
    \item \textbf{Soft light:} Soft-edged shadows; natural shadow transitions or nearly imperceptible shadows; low contrast; relatively rich shadow detail; weak directionality; typically from a large source (e.g., a candle with a lampshade enlarging and softening the source). Common sources: overcast light, softbox, snow light.

    The \textbf{Key checks} include does the image show low contrast and soft shadow transitions?

    \item \textbf{Hard light:} Hard-edged shadows with very clear contours; strong contrast with large differences between light and dark; typically comes from a concentrated light source (in mixed lighting, there is a visible or implied concentrated-source effect). Common sources include direct sunlight, a bare bulb, a spotlight, or firelight close to the subject.

    The \textbf{Key checks} include does the image show high contrast with sharp, crisp shadows?

    \textbf{Note:} First determine whether the prompt requires soft or hard light, then judge whether the video satisfies the aesthetic standard under that requirement.
\end{enumerate}

\textbf{C6. Light direction.} 

\begin{enumerate}
    \item \textbf{Top light:} Source directly above the subject, shining downward. Often creates strong facial shadows (eye sockets, under nose, chin), producing a skull-like or dramatic effect; used for oppressive, mysterious, tense, or exaggerated moods (common in horror/thriller), also for special styling or stage lighting.

     \item \textbf{Butterfly light (Paramount):} Light facing the subject from slightly above, creating a butterfly-shaped shadow under the nose. Soft yet sculpted; suitable for refined facial features; common in fashion/beauty and feminine portraits for an elegant, noble look.
     
     \item \textbf{Front light:} Light direction aligned or nearly aligned with the camera axis. Produces even, flat lighting; reproduces inherent object colors well.
     
     \item \textbf{Front-side light (three-quarter front / front sidelight):} About 45 degrees off the camera axis. Produces a large lit area with some shadows; adds depth and layering; reveals texture well.
     
     \item \textbf{Side light:} About 90 degrees to the camera axis. Strong contrast; depth, layering, and texture are pronounced; highly expressive.
     
     \item \textbf{Back-side light (three-quarter back / 135 degrees):} About 135 degrees to the camera axis. Smaller lit area and larger shadow area; strong modeling and outline. For closer portraits, some fill is often added to keep a reasonable light ratio between lit and backlit sides.
     
     \item \textbf{Rembrandt light:} Light from above and to one side, creating a triangular highlight on the cheek (usually beside the nose), with the opposite side in shadow. Strong depth and drama; emphasizes facial structure; common in portraits or serious, profound character looks.
     
     \item \textbf{Loop lighting:} Light slightly off-center in front (typically 30 degrees to 45 degrees), creating a small soft shadow looping from the nose toward the cheek.
     
     \item \textbf{Split lighting:} Divides the face into two halves: one fully lit, the other in shadow, forming strong contrast.
     
     \item \textbf{Under light (uplight):} Source below the subject shining upward. Creates unnatural or eerie shadows on the face/chin/body; used for horror, mystery, unease, or dramatic moods.
     
     \item \textbf{Rim light:} Contour light from behind or behind-side to outline the subject, enhancing depth and separation from the background (especially when subject/background colors are similar). Often brighter than the key light; commonly achieved with backlight or back-side light.
     
     \item \textbf{Backlight:} Source behind the subject shining toward the camera from the rear. Used to outline the subject, enhance depth/layers, and separate subject from background; widely used to create mood/emotion and visual beauty (dreamy, romantic, mysterious).
     
     \item \textbf{Counterlight:} Light placed behind the subject, shining from the back toward the camera; can create clear outlines, silhouettes, or special effects (transmitted light, glints) to enhance layering and translucency.
     
     \item \textbf{Silhouette:} Shot in strong backlight such that the subject becomes a dark outline while the background is very bright, emphasizing shape/pose with strong visual contrast.
     
     \textbf{Note:} First determine which lighting direction the prompt asks for, then verify whether the image's lighting direction matches the corresponding characteristics. A good image must satisfy the prompt's lighting-direction requirement and the associated lighting traits.
    
\end{enumerate}

\subsection{Visual Fidelity (VF)}
\label{app:vf}
The video should look more realistic and closer to the real world. Judge primarily based on the following visual aspects.

\textbf{C7. Interaction Fidelity.} Evaluate whether interactions between objects, and between objects and the environment, conform to physical common sense and spatial logic.  The \textbf{Key checks} include
\begin{enumerate}
    \item \textbf{Collisions and contact:} Do collisions show reasonable physical feedback (e.g., bouncing, deformation)? Is there any illogical interpenetration between objects (i.e., clipping)?
    \item \textbf{Occlusion relationships:} Is the front and back spatial ordering correct? Can foreground objects accurately and stably occlude background objects?
    \item \textbf{Contact-point accuracy:} Are contact points between characters/objects and planar surfaces (ground, tabletop, etc.) stable and correct? (e.g., feet floating above the ground; handheld objects not fitting tightly in the palm)
\end{enumerate}

\textbf{C8. Physical Adherence.} Evaluate whether dynamic effects follow basic real-world physics such as gravity, inertia, and fluid dynamics. The \textbf{Key checks} include
\begin{enumerate}
   \item \textbf{Gravity and motion:} Are falling motions and projectile trajectories natural? Do motion start-up, acceleration, deceleration, and stopping follow inertia?
    \item \textbf{Fluid simulation:} Are dynamics of fluids such as water, smoke, fire, and clouds (e.g., flow, surface ripples, diffusion, combustion shapes) realistic and natural?
    \item \textbf{Material dynamics:} Do flexible materials (e.g., cloth, hair) swing and flutter in ways consistent with their material properties and external forces? Are rigid-body breakage/fracture effects believable?
\end{enumerate}

\textbf{C9. Structural Stability.} Evaluate whether the main subject and background environment exhibit unreasonable shaking, flicker, deformation, or distortion. The \textbf{Key checks} include
\begin{enumerate}
\item \textbf{Shape stability:} Does the subject (especially people or animals) maintain consistent structure? Are there abnormal twists, stretches, or unnatural melting in limbs or facial features?
\item \textbf{Background/texture stability:} Are background elements or surface textures stable? Is there unexplained jitter, flicker, or pattern changes over time?
\item \textbf{Temporal continuity:} Does the same object keep consistent identity and attributes across consecutive frames, avoiding sudden popping changes or morphing into another object?
  \end{enumerate}

\textbf{C10. Sharpness.} Evaluate sharpness and the ability to render fine details.
  \begin{enumerate}
      \item \textbf{Sharpness:} Are edges and contours crisp and well-defined, or is the image generally soft/blurred with insufficient detail acuity?
      \item \textbf{Detail textures:} Are surface textures (e.g., skin pores, fabric fibers, wood grain) clearly visible, or is there obvious smudging/painterly look that causes major detail loss?
      \item \textbf{Artifacts and noise:} Are there unnatural color blocks, mosaic/compression artifacts, or abnormal digital noise that harms overall visual quality?
  \end{enumerate}

\subsection{Visual Plausibility (VP)}
\label{app:vp}
The visual content should be more reasonable in terms of shot size, viewpoint, and richness of details. You must first combine the prompt requirements with the on-screen content to determine whether the video meets the frame-reasonableness criteria.

\textbf{C11. Shot size.} Shot size refers to the distance relationship between the camera and the subject. Categories and definitions:

\begin{enumerate}
    \item \textbf{Extreme long shot:} A very vast scene; the character occupies a very small portion of the frame; typically used to show the grand environment

     \item \textbf{Long shot:} Balances the subject and surrounding scenery; usually includes the subject's full body while also establishing the environment.
     \item \textbf{Full shot:} The subject fills the composition from head to toe, with some environment included.
     \item \textbf{Medium shot:} Frames a person from the waist up.
     \item \textbf{Medium close-up:} Composition is roughly up to the chest; often emphasizes the face while still retaining some distance from the subject.
     \item \textbf{Medium full shot:} The main character is shown to around the waist; the surrounding environment is fully established.
     \item \textbf{Close-up:} Frames a person from the chest up.
     \item \textbf{Close shot:} Only a part of the subject is filmed.
     \item \textbf{Extreme close-up:} Only a very small part of the subject is shown (a tiny facial area or a detailed part of an object).
     
     \textbf{Note:} Identify the shot-size requirement in the prompt, then judge whether the video matches the defining characteristics. For example, if the prompt asks for an extreme close-up and the video shows an extreme close-up of a character's hand, it can be considered compliant.
\end{enumerate}

\textbf{C12. Shot Composition.} Composition categories and definitions:
\begin{enumerate}
    \item \textbf{Centered composition:} The subject is placed at the exact center of the frame.
    \item \textbf{Rule-of-thirds composition:} Divide the frame into thirds horizontally and vertically, forming four intersection points; the subject is placed on one of these intersections.
    \item \textbf{Left-weighted composition:} Major visual elements concentrate on the left; the right side is relatively empty or visually lighter.
    \item \textbf{Right-weighted composition:} Major visual elements concentrate on the right; the left side is relatively empty or visually lighter.
    \item \textbf{Balanced composition:} The visual weight of different elements is coordinated and stable across the frame.
    \item \textbf{Symmetrical composition:} Left and right (or top--bottom) symmetry; multiple identical/similar subjects arranged with axial or central symmetry.
    \item \textbf{Diagonal composition:} The subject or key elements are arranged along a diagonal direction.
    \item \textbf{Leading-lines composition:} Uses line elements in the frame to actively guide the viewer's gaze toward the subject or into image depth.
    \item \textbf{Frame-within-a-frame composition:} Foreground objects form a picture frame that encloses the subject.
    \item \textbf{Edge composition:} The subject is placed tightly against the left/right/top/bottom edge of the frame.
    \item \textbf{Negative-space composition:} The subject occupies a small area and is surrounded by large empty/negative space.
    \item \textbf{Triangular composition:} Three subjects (or key points) form a triangular layout in the frame.

    \textbf{Note:} Identify the prompt's composition requirement, then judge whether the video matches the definition. For example, if the prompt requires a centered composition but the subject is actually at the edge, the video does not meet the requirement.
\end{enumerate}

\textbf{C13. Focal length.} Lens focal-length categories and definitions:

  \begin{enumerate}
  \item \textbf{Ultra-wide:} Very wide field of view; captures a huge area. For example, horizons/railings are stretched and distorted due to perspective; the coverage is extremely wide.
  \item \textbf{Wide-angle:} More natural than ultra-wide while still wide. Foreground--background perspective is present but not overly exaggerated.
  \item \textbf{Standard / short portrait focal length:} Field of view close to human vision; looks natural. Background is slightly blurred but not very strong.
  \item \textbf{Medium telephoto / portrait focal length:} Tighter framing; good for half-body or portrait close-ups. Strong background compression; shallow depth of field; background blurs easily.
  \item \textbf{Telephoto:} Pulls distant subjects closer; narrow field of view. Obvious background blur; strong compression. Background can be roughly recognized but looks quite soft.
  \item \textbf{Super-telephoto:} Captures very distant subjects; extremely narrow field of view. Distant scenes are magnified and become extremely blurred; the background may be unrecognizable.
  \item \textbf{Fisheye:} Extremely wide view (up to 180 degrees or more) with obvious curved distortion; the image looks spherical, with exaggerated bending near edges. Common in peephole/security-camera-like views where straight horizontal/vertical lines bend into globe-like arcs.
  \item \textbf{Macro lens:} Focuses at very close distances; captures tiny-object details and textures not easily visible to the naked eye.
  \item \textbf{Tilt-shift lens:} \textbf{Tilt} changes the focal-plane direction so objects at different distances can be sharp simultaneously or only a local area is sharp. \textbf{Shift} corrects perspective distortion. Typically: center region sharp with surrounding blur.
  
  \textbf{Note:} Determine the focal-length requirement from the prompt, then judge whether the video matches the definition. For example, if the prompt requires a fisheye lens but there is no obvious curved distortion, it is non-compliant.
  \end{enumerate}

\textbf{C14. Camera Angle.} 
\begin{enumerate}

  \item \textbf{Eye-level:} Camera is on the same horizontal level as the subject.
  \item \textbf{Low angle (upward-looking):} Camera shoots from below upward.
  \item \textbf{High angle (downward-looking):} Camera shoots from above downward.
  \item \textbf{Bird's-eye (aerial top-down):} An extreme high angle; camera is almost perpendicular to the ground, shooting straight down.
  \item \textbf{Extreme low angle:} Shoots upward from a very low position.
  \item \textbf{Dutch angle (tilted):} Camera is tilted relative to the subject.
  \item \textbf{Over-the-shoulder:} Camera is behind one character while facing another; the former's shoulder/back is visible to the audience.
  
  \textbf{Note:} Identify the prompt's shooting-angle requirement, then judge compliance based on the definition. For example, if the prompt asks for a bird's-eye view and the frame clearly shows a top-down scene, then it is compliant.
\end{enumerate}

\textbf{C15. Detail Richness.} This dimension evaluates how filled the frame is and whether the background environment is logically consistent with the subject.

\begin{enumerate}
    \item \textbf{Detail Richness:} The content is sufficiently rich and appropriate for the shot size and theme. In close-ups, textures and materials are clear; in long shots, environmental elements (architecture, vegetation, crowds) are plentiful and reasonably organized.  
    
    A bad case is when the frame feels empty or monotonous and lacks the expected level of detail—for example, a bustling metropolis with only a few blurry building silhouettes, or a close-up of exquisite food where ingredient textures aren’t discernible.

    \item \textbf{Environmental coherence:} Background/props/lighting and other environmental elements are logically consistent with the subject, era, and narrative context (e.g., no modern objects in an ancient setting; a sci-fi setting has a unified design language).  
    
    Continuity errors or illogical elements are present. For example, in a scene depicting a desolate desert, an oasis or modern architecture appears in the background where it does not belong, undermining the realism and immersiveness of the frame.
\end{enumerate}

The three core dimensions and fifteen criteria above span color quality, camera angles, and lighting/shadow, providing a systematic framework for evaluating video aesthetics.

\section{Expert-Level Video Aesthetic Preference Data Collection}
\phantomsection
\label{app:Expert-Leve}

Our training data and the benchmark in \cref{sec:data} were annotated by six experts from professional art academies. Following our 15 evaluation criteria, each annotator conducted pairwise comparisons along three dimensions: VA, VF, and VP. For any pair (Video A, Video B), each dimension was labeled as $1$ if $A \succ B$, $-1$ if $A \prec B$, and $0$ for a tie. Each sample was evaluated by two independent annotators; in cases of disagreement, a third expert reviewed the sample, and the final label was determined by majority vote. This pipeline produced approximately 20000 high-quality samples, which were split into a training set for AesRM and an evaluation set, AesVideo-Bench. AesVideo-Bench contains 2639 samples. Using the aggregate score over VA, VP, and VF (i.e., summing the three dimension labels), the overall preference distribution is well-balanced: $A \succ B$ (47.78\%), $A \prec B$ (51.27\%), and ties (0.95\%). 



\section{Three-Stage Training of AesRM}
\phantomsection
\label{app:Three-Stage}

\subsection{Stage 1: Atomic Aesthetic Capability Learning}
\phantomsection
\label{app:Atomic-Aesthetic-Capability-Learning}

Stage 1 uses SFT to improve the base model’s ability to predict atomic aesthetic attributes, where the base model is required to output the categories of 16 atomic aesthetic concepts for a given image, and human expert annotations serve as the ground truth. Specifically,

\begin{table*}[t!]
\centering
\caption{The prompt for the atomic aesthetic capability learning stage.}
\vspace{-0.1mm}
\label{tab:sft_prompt_atomic_aesthetics}
\begin{tabular}{p{0.96\textwidth}}
\hline
\textbf{Prompt Text} \\
\hline
You are a film frame analysis assistant. Please identify visual attributes based on the given image. You must output \textbf{only one JSON object} and do not output any additional text. The JSON must contain the following keys (options in parentheses):
\begin{itemize}
  \item \textbf{Light Style} (\texttt{Practical Light}, \texttt{Motivated/Scene Light}, \texttt{None})
  \item \textbf{Light Source} (\texttt{Daylight}, \texttt{Sunny Daylight}, \texttt{Overcast Daylight}, \texttt{Moonlight}, \texttt{Snow Light}, \texttt{Firelight}, \texttt{Fluorescent}, \texttt{Incandescent}, \texttt{Neon}, \texttt{LED}, \texttt{High-intensity Light}, \texttt{Artificial Light}, \texttt{Mixed Light}, \texttt{None})
  \item \textbf{Light Quality} (\texttt{Soft Light}, \texttt{Hard Light}, \texttt{None})
  \item \textbf{Light Direction} (\texttt{Front Light}, \texttt{Front-side Light}, \texttt{Side Light}, \texttt{Rembrandt Light}, \texttt{Ring Light}, \texttt{Split Light}, \texttt{Side Back Light}, \texttt{Back Light}, \texttt{Rim Light}, \texttt{Silhouette}, \texttt{Top Light}, \texttt{Butterfly Light}, \texttt{Under Light}, \texttt{None})
  \item \textbf{Color Contrast} (\texttt{High Contrast}, \texttt{Medium Contrast}, \texttt{Low Contrast})
  \item \textbf{Lighting Quality} (\texttt{Good}, \texttt{Medium}, \texttt{Poor})
  \item \textbf{Color Palette} (\texttt{Warm Tone}, \texttt{Cool Tone}, \texttt{Mixed Tone}, \texttt{Monochrome}, \texttt{Neutral/Normal Tone}, \texttt{None})
  \item \textbf{Saturation} (\texttt{Low Saturation}, \texttt{Saturation: Slightly Low/Normal}, \texttt{Saturation: Normal}, \texttt{High Saturation}, \texttt{Over-saturated}, \texttt{None})
  \item \textbf{Exposure} (\texttt{Brightness: Too Dark}, \texttt{Brightness: Slightly Dark}, \texttt{Brightness: Normal}, \texttt{Brightness: Slightly Bright}, \texttt{Brightness: Overexposed}, \texttt{None})
  \item \textbf{Sharpness} (\texttt{Sharpness: Blurry}, \texttt{Sharpness: Slightly Soft}, \texttt{Sharpness: Clear}, \texttt{Sharpness: Very Clear}, \texttt{None})
  \item \textbf{Shot Composition} (\texttt{Centered Composition}, \texttt{Balanced Composition}, \texttt{Right-weighted Composition}, \texttt{Left-weighted Composition}, \texttt{Symmetrical Composition}, \texttt{Rule of Thirds}, \texttt{Diagonal Composition}, \texttt{Leading Lines}, \texttt{Framing Composition}, \texttt{Edge Composition}, \texttt{Triangular Composition}, \texttt{Negative Space Composition}, \texttt{None})
  \item \textbf{Focal Length} (\texttt{Super Telephoto}, \texttt{Telephoto}, \texttt{Normal}, \texttt{Wide-angle}, \texttt{Ultra Wide-angle}, \texttt{Fisheye}, \texttt{Macro}, \texttt{Tilt-shift}, \texttt{None})
  \item \textbf{Shot Size} (\texttt{Extreme Close-up}, \texttt{Close-up}, \texttt{Medium Close-up}, \texttt{Medium Shot}, \texttt{Medium Long Shot}, \texttt{Long Shot}, \texttt{Extreme Long Shot}, \texttt{None})
  \item \textbf{Lens Type} (\texttt{Clean Single-person Shot}, \texttt{Two-person Shot}, \texttt{Group Shot}, \texttt{Three-person Shot}, \texttt{Establishing Shot}, \texttt{Insert Shot}, \texttt{None})
  \item \textbf{Camera Angle} (\texttt{Eye-level}, \texttt{Over-the-shoulder}, \texttt{Low Angle}, \texttt{Extreme Low Angle}, \texttt{High Angle}, \texttt{Top-down}, \texttt{Aerial}, \texttt{Dutch Angle}, \texttt{None})
  \item \textbf{Composition Quality} (\texttt{Good}, \texttt{Medium}, \texttt{Poor})
\end{itemize}
\textbf{Output Example:} \texttt{\{"Light Style":"None","Light Source":"LED" \ldots\}}\\
\hline
\end{tabular}
\end{table*}

\noindent \textbf{Training Prompt} of Stage 1 is shown in \cref{tab:sft_prompt_atomic_aesthetics} where the base model is asked to predict 16 basic atomic aesthetic concepts, and we provide a corresponding set of options for each concept to improve the accuracy of the model’s responses.

\noindent \textbf{Training Details.} Using InternVL 3.5 as the backbone, we perform full-parameter SFT on an internal dataset of about 15K atomic aesthetic samples, guided by the system prompt in \cref{tab:sft_prompt_atomic_aesthetics}. The model is trained for 3 epochs with a learning rate of $1 \times 10^{-5}$, a 0.05 warmup ratio, and an effective batch size of 16 (8 per device with 2 accumulation steps). The process is completed in approximately 5 hours on a cluster of 8 NVIDIA A100 GPUs.


\subsection{Stage 2: Cold-Start for Aesthetic Evaluation}
\phantomsection
\label{app:Cold-Start-for-Aesthetic-Evaluation}

\noindent \textbf{Self-Consistency-based CoT Synthesis.} To synthesize high-quality CoT data and mitigate hallucinations in the teacher model, i.e., Gemini 2.5 Pro, we employ the Self-Consistency-based CoT Synthesis proposed in \cref{sec:stage2}. Specifically, we randomly select over 3k samples from the collected preference data in \cref{sec:data_collect}. Guided by \texttt{Expert Reason}, we generate CoT explanations encompassing 15 fine-grained criteria, performing $M=4$ samplings per sample to solve \cref{eq:cot}. Notably, given the finite set of 15 criteria, we treat the selection probability $\alpha_{j,i}$ as a hard label (i.e., 0 or 1). We then exhaustively enumerate all combinations to maximize the cumulative consistency score across all criteria, while ensuring the final three-dimensional labels strictly align with the \texttt{Expert Label}. This process yields a final dataset of 3958 paired samples, each including 15-criteria CoT across VA, VF and VP and corresponding final labels.

\noindent \textbf{Training Prompt.} The Stage-2 system prompts for AesRM-Base and AesRM-CoT are as follows.

\small
\setlength{\LTpre}{0pt}
\setlength{\LTpost}{0pt}
\setlength{\LTleft}{0pt}
\setlength{\LTright}{0pt}
\renewcommand{\arraystretch}{1.12}
\setlength{\emergencystretch}{2em}

\setlist[itemize]{leftmargin=1.2em,itemsep=0.15em,topsep=0.2em,parsep=0pt}
\setlist[enumerate]{leftmargin=1.6em,itemsep=0.15em,topsep=0.2em,parsep=0pt}

\begin{longtable}{@{\hspace{0.6em}}p{0.96\linewidth}@{\hspace{0.6em}}}
\caption{Full system prompt for AesRM-Base.}
\label{tab:full_prompt_aesrm_base}\\
\toprule
\textbf{Prompt Text} \\
\midrule
\endfirsthead

\toprule
\textbf{Prompt Text (continued)} \\
\midrule
\endhead

\endfoot

\bottomrule
\endlastfoot

You are a seasoned film and television analyst with a rigorous, detail-oriented approach and deep expertise in video aesthetics. You are also an accomplished copywriter. You will be given two videos: Video A and Video B. Both videos are generated from the same prompt.

\vspace{0.35em}
\textbf{Input}\\
\begin{itemize}
  \item \textbf{Video A:} The video object will be provided immediately following the text label \texttt{Video A}.
  \item \textbf{Video B:} The video object will be provided immediately following the text label \texttt{Video B}.
  \item A shared video generation prompt applies to both videos.
\end{itemize}

\vspace{0.35em}
\textbf{Mandatory compliance}\\
Review Video A, Video B, and the prompt. If you are unable to view the videos or access the prompt, you must immediately terminate the analysis, skip all tasks below, and must not fabricate any analysis based solely on the prompt.

\vspace{0.35em}
\textbf{Evaluation dimensions}\\
You are required to comprehensively assess the relative performance of Video A and Video B across 3 dimensions and 15 criteria.

\vspace{0.2em}
\textit{Refer to \cref{app:Hierarchical} for detailed explanations of the three core dimensions and the 15 criteria.}

\vspace{0.35em}
\textbf{Output format (must be followed strictly)}\\
\begin{enumerate}
    \item Output only one sentence as the final evaluation, covering the comparison between Video A and Video B across all three dimensions, with no analysis or reasoning.
    
    \item For each dimension, the comparison must be one of: Video A outperforms Video B / Video A underperforms Video B / the two are comparable. For example: Video A underperforms Video B in visual aesthetics, while the two are comparable in visual fidelity and visual plausibility.
    
    \item Wrap the conclusion in \texttt{<answer>Your evaluation</answer>}.

\end{enumerate}
\end{longtable}


\small
\setlength{\LTpre}{0pt}
\setlength{\LTpost}{0pt}
\setlength{\LTleft}{0pt}
\setlength{\LTright}{0pt}
\renewcommand{\arraystretch}{1.12}
\setlength{\emergencystretch}{2em}

\setlist[itemize]{leftmargin=1.2em,itemsep=0.15em,topsep=0.2em,parsep=0pt}
\setlist[enumerate]{leftmargin=1.6em,itemsep=0.15em,topsep=0.2em,parsep=0pt}

\begin{longtable}{@{\hspace{0.6em}}p{0.96\linewidth}@{\hspace{0.6em}}}
\caption{Full system prompt for AesRM-CoT.}
\label{tab:full_prompt_aesrm_cot}\\
\toprule
\textbf{Prompt Text} \\
\midrule
\endfirsthead

\toprule
\textbf{Prompt Text (continued)} \\
\midrule
\endhead

\endfoot

\bottomrule
\endlastfoot

You are a seasoned film and television analyst with a rigorous, detail-oriented approach and deep expertise in video aesthetics. You are also an accomplished copywriter. You will be given two videos: Video A and Video B. Both videos are generated from the same prompt.

\vspace{0.35em}
\textbf{Input}\\
\begin{itemize}
  \item \textbf{Video A:} The video object will be provided immediately following the text label \texttt{Video A}.
  \item \textbf{Video B:} The video object will be provided immediately following the text label \texttt{Video B}.
  \item A shared video generation prompt applies to both videos.
\end{itemize}

\vspace{0.35em}
\textbf{Mandatory compliance}\\
Review Video A, Video B, and the prompt. If you are unable to view the videos or access the prompt, you must immediately terminate the analysis, skip all tasks below, and must not fabricate any analysis based solely on the prompt.

\vspace{0.35em}
\textbf{Evaluation dimensions}\\
You are required to comprehensively assess the relative performance of Video A and Video B across 3 dimensions and 15 criteria.

\vspace{0.2em}
\textit{Refer to \cref{app:Hierarchical} for detailed explanations of the three core dimensions and the 15 criteria.}

\vspace{0.35em}
\textbf{Reasoning requirements (must be followed strictly)}\\
In \verb|<think>|, include your reasoning process and write it strictly in the following order and format, criterion by criterion:

\begin{itemize}
  \item Order: (1) \textbf{Visual Aesthetics} (C1--C6) $\rightarrow$
        (2) \textbf{Visual Fidelity} (C7--C10) $\rightarrow$
        (3) \textbf{Visual Plausibility} (C11--C15).
  \item Each criterion must start with a fixed tag line in the format
        \verb|[Dimension-CriterionLetter.CriterionName]|, e.g.,
        \verb|[Visual Aesthetics-C1.Color Quality]|.
        The dimension name must be one of \texttt{Visual Aesthetics}, \texttt{Visual Fidelity}, or \texttt{Visual Plausibility}.
        The criterion letter and name must match the original definition.
  \item For each criterion, write \textbf{exactly four lines} in the following fixed format:
    \begin{enumerate}
      \item The prompt’s requirement for this criterion and the corresponding standard (use professional terminology and the criterion definition).
      \item Whether Video A satisfies the requirement and why.
      \item Whether Video B satisfies the requirement and why.
      \item \textbf{Score (A relative to B):} \{1, -1, 0\}, with a brief justification.
    \end{enumerate}
  \item \textbf{Scoring rule:} If Video A is better than Video B (with clear evidence), assign 1; if worse, assign -1; if similar / evidence is insufficient / not applicable, assign 0.
  \item \textbf{Severe violation rule:} If a severe violation occurs in any dimension (e.g., obvious clipping/interpenetration, broken limbs, major physics violation, severe stability collapse, extreme overexposure or underexposure), explicitly point it out under the corresponding criterion. In the final decision for that dimension, directly mark the affected video as disadvantaged (A disadvantaged if the violation occurs in A; otherwise B).
  \item Sum the scores across all criteria within each dimension to obtain the dimension score.
\end{itemize}

\vspace{0.35em}
\textbf{Example}\\
{\ttfamily
\setlength{\parindent}{0pt}
[Visual Aesthetics-C2.Time Period]\par
- The prompt requires dusk. Dusk standard: tones lean golden/orange; the sky gradually transitions into night; cool colors emerge while warm colors fade.\par
- Video A: meets dusk characteristics; contains orange-gold elements; lighting is dim and gradually transitions toward night.\par
- Video B: does not meet dusk characteristics; lacks orange-gold elements; sky is overly dark with lost detail.\par
- Score (A v.s. B): 1 (A is clearly better)\par

[Summary of Visual Aesthetics: In Visual Aesthetics, the sum over 6 criteria is -1+1+1+1+1+1=4>0. Therefore, the Visual Aesthetics score is positive, and Video A is better than Video B in Visual Aesthetics.]
}

\vspace{0.35em}
\textbf{Output format (must be followed strictly)}\\
Your output must follow this order: first \verb|<think> your reasoning process </think>|, then \verb|<answer> a one-sentence final evaluation </answer>|.

\end{longtable}

\noindent \textbf{Training Details.} We use atomic-enhanced InternVL 3.5 and perform full-parameter SFT on 3958 synthesized CoT samples to teach AesRM pairwise aesthetic reasoning and comparison. We train for 6 epochs with a learning rate of \(1\times10^{-5}\), a warmup ratio of 0.05, and an effective batch size of 4 (4 per device with 4 accumulation steps). Training finishes in about 9 hours on 4 NVIDIA A100 GPUs.


\subsection{Stage 3: GRPO for Aesthetic Evaluation}
\phantomsection
\label{app:GRPO-for-Aesthetic-Evaluation}

\noindent \textbf{Rewards for AesRM-CoT}.  In this section, we discuss the reward design details and hyperparameter settings for AesRM-Base and AesRM-CoT during the GRPO stage.

For \textbf{AesRM-Base}, we use two rewards: a format reward \(R_{\text{fmt}}\) and an accuracy reward \(R_{\text{acc}}\). \(R_{\text{fmt}}\) enforces that the model compares the video pair along VA, VF, and VP and wraps the final conclusion in \texttt{<answer>}. \(R_{\text{acc}}\) encourages the predicted labels to match the ground truth (see \cref{eq:acc_reward}). Specifically, we map each dimension-wise match signal (VA/VF/VP) to \([-1,1]\) via
    \begin{equation}
        S_{\text{norm}} = (S_{\text{original}}- 0.5) \times 2,
    \end{equation}
    and compute a weighted sum across the three dimensions, with weights \(0.3\), \(0.2\), and \(0.5\), respectively, to strengthen supervision on VP where AesRM-Base is relatively weak (See the discussion in \cref{app:More-Results}. The reward weight $\lambda$ in \cref{eq:acc_reward} is $0.1$).

\begin{figure}[t!]
\centering
  \includegraphics[width=1\textwidth, height=0.22\textheight]{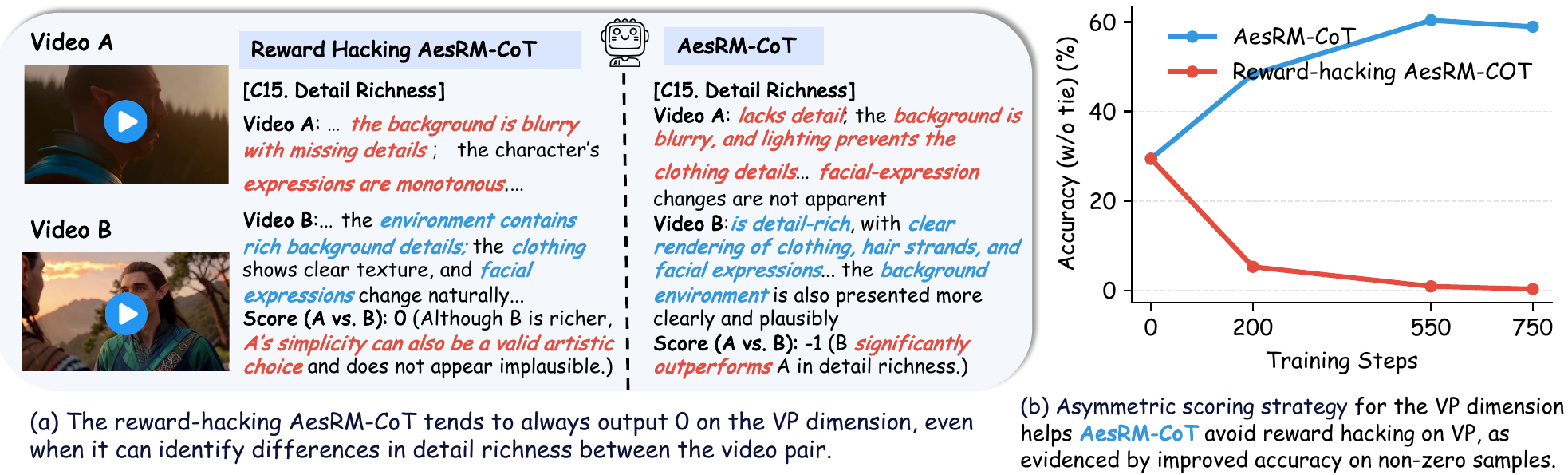}
\caption{(a) Due to the sparsity of non-zero samples in the VP dimension, AesRM-CoT is prone to reward hacking: it may correctly identify differences between the video pair (e.g., Video B has richer details) but still outputs a score of 0. (b) We therefore design an asymmetric scoring strategy: for criteria where the teacher assigns a non-zero score, the student is encouraged to output a non-zero score rather than 0 through asymmetric rewards for non-zero vs. zero predictions. This strategy improves AesRM-CoT’s accuracy on non-zero VP samples and substantially alleviates score collapse.}
\label{fig:vp_reward_hacking}
\end{figure}
For \textbf{AesRM-CoT},  rewards include
\begin{itemize}
    \item The format reward \(R_{\text{fmt}}\) implements a hierarchical validation mechanism. Initially, it mandates a structured output following a strict \texttt{<think>} then \texttt{<answer>} sequence and failure to adhere to this format results in a zero reward. Upon successful structural validation, the system further analyzes the CoT to verify the inclusion of 15 pre-defined fine-grained aesthetic criteria. Specifically, the reward function employs regular expressions to detect the presence of these 15 criteria: each identified criterion earns 1 point, otherwise 0. The cumulative score is finally summed and normalized to a range of $[-1, 1]$.
    \item The accuracy reward \(R_{\text{acc}}\) is the same as in AesRM-Base, measuring the discrepancy between the model’s final reasoning outcome and the ground-truth labels.
    \item The consistency reward, $R_{\text{cst}}$, comprises both internal and external consistency checks. Specifically, internal consistency requires parsing the \texttt{<think>} section to extract individual scores ($-1, 0, 1$) assigned to fine-grained criteria within each dimension—for instance, extracting a score of 1 from \texttt{Score (A v.s. B): 1} as shown in \cref{tab:full_prompt_aesrm_cot}. The reward function then verifies whether the algebraic sum of these criteria-level scores aligns with the final dimension-level conclusion presented in the summary paragraph of that dimension (e.g., the score of 4 in the \texttt{Summary of Visual Aesthetics}). External consistency, conversely, evaluates the alignment between the dimension-level conclusions derived within the \texttt{<think>} section and the final labels provided in the \texttt{<answer>}. For each dimension VA, VP and VF, a score of 1 is awarded if and only if both internal and external consistency are simultaneously satisfied; otherwise, the dimension receives a score of 0. Finally, these scores are aggregated across the three dimensions and normalized to the range of $[-1, 1]$ to yield the final $R_{\text{cst}}$.
    \item The process reward \(R_{\text{prc}}\) is designed to compare the student model (AesRM-CoT) against the teacher model (Gemini 2.5 Pro) at the criterion level by measuring CoT similarity. As shown in \cref{eq:erward_prc}, we quantify student–teacher alignment using both semantic similarity and lexical overlap. Importantly, we compute this similarity only when the student and teacher assign the same per-criterion score, i.e., the extracted value from \texttt{Score (A v.s. B): 1/-1/0} is identical (see \cref{tab:full_prompt_aesrm_cot}). We then aggregate similarity scores over all 15 criteria and normalize the final reward to \([-1,1]\).
    
    In practice, we observe reward hacking in the VP dimension due to label imbalance in the GRPO data: a large fraction (78.23\%) of samples have a VP label of 0 (i.e., no difference between Video A and Video B), which can cause the model’s dimension-wise predictions to collapse to 0. As illustrated in \cref{fig:vp_reward_hacking}(a), the model may correctly detect differences (e.g., in detail richness) but still prefer outputting 0. To mitigate this, we adopt an \textbf{asymmetric scoring strategy} that introduces a VP-specific shaping term in \(R_{\text{prc}}\): when the teacher predicts a non-zero score (\(\pm 1\)) and the student also predicts a non-zero score, the student receives a partial weight of 0.5 even if the signs do not match. Concretely, in \cref{eq:erward_prc}, the indicator \(\mathbb{I}\!\left(s_{(i)}^{\text{teacher}}=s_{(i)}^{\text{student}}\right)\) is replaced by a weight of 0.5 when \(\,s_{(i)}^{\text{teacher}}\neq s_{(i)}^{\text{student}}\) but \(s_{(i)}^{\text{teacher}}\neq 0\) and \(s_{(i)}^{\text{student}}\neq 0\). \cref{fig:vp_reward_hacking}(b) shows this strategy encourages AesRM-CoT to explore non-zero outputs and prevents degenerate collapse to all-zero predictions when non-zero cases are sparse .
\end{itemize}

\noindent \textbf{Training Prompt.} During the GRPO stage, we train both AesRM-Base and AesRM-CoT using the same system prompts as in the cold-start stage (\cref{tab:full_prompt_aesrm_base} and \cref{tab:full_prompt_aesrm_cot}).

\begin{figure}[t!]
\centering
  \includegraphics[width=1\textwidth, height=0.2\textheight]{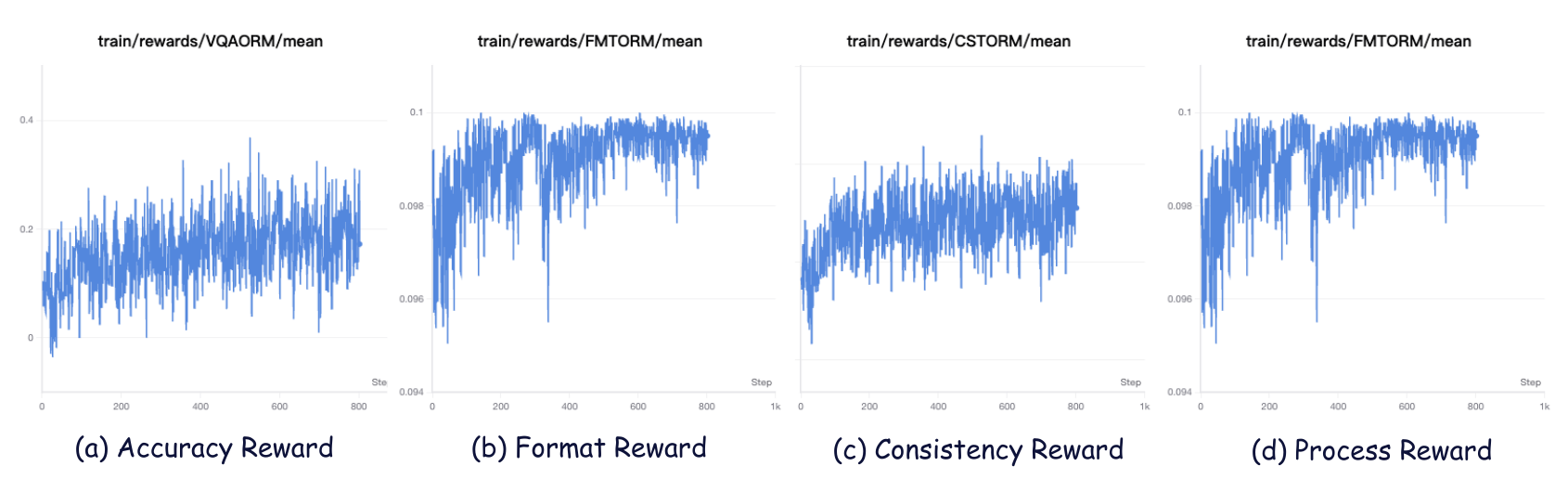}
\caption{AesRM-CoT’s rewards steadily increase throughout the GRPO stage.}
\label{fig:training_rewards}
\end{figure}

\noindent \textbf{Training Details.} During the GRPO stage, we train on 16566 samples. We set the reward weights \(\lambda\) for \(R_{\text{acc}}, R_{\text{fmt}}, R_{\text{cst}},\) and \(R_{\text{prc}}\) to 1, 0.1, 0.5, and 1, respectively. For \(R_{\text{acc}}\), the three dimension weights remain 0.3, 0.2, and 0.5. We use a learning rate of \(5\times10^{-7}\) and train for 5 epochs, with default settings \(\beta=0.1\), temperature 1, \(top_k=50\), and \(top_p=0.9\). Training is conducted on 32 A100 GPUs and takes approximately three days. \cref{fig:training_rewards} shows that AesRM-CoT’s rewards steadily increase throughout the GRPO stage. In particular, the rise in process reward highlights the effectiveness of our CoT-similarity-based reward signal.

\section{Experiments}
\phantomsection
\label{app:exps}
In this section, we provide additional training details and present further quantitative and qualitative results.

\subsection{Evaluation Benchmarks and Metrics}
\phantomsection
\label{app:Evaluation}
In addition to AesVideo-Bench, we evaluate on three additional benchmarks in \cref{sec:exp}, as follows.

\textbf{VideoReward-Bench} \citep{liu2025improving} is a pairwise video preference dataset collected from the Internet with diverse prompts. The prompts are grouped into eight categories, animals, architecture, food, humans, plants, scenes, vehicles, and objects, and are further expanded using GPT-4o \citep{openai2024gpt4technicalreport}. Professional annotators conduct pairwise comparisons along three axes, Visual Quality (VQ), Motion Quality (MQ), and Text Alignment (TA), labeling each video pair as \emph{A wins} / \emph{tie} / \emph{B wins}. The dataset contains approximately 1.3K video pairs with three-dimensional label vectors. In our experiments, we focus on the Visual Quality dimension, which is most relevant to video aesthetics. It is worth noting that VideoReward-Bench follows the same data distribution as the baseline VideoAlign \citep{liu2025improving} training data described in \cref{sec:exp-rm}.

\textbf{GenAI-Bench} \citep{jiang2024genai} consists of approximately 1900 short video pairs (2s to 2.5s in duration), aggregating data from six pre-SOTA and four recent open-source T2V models. Preferences are determined through user voting on the GenAI-Arena platform \citep{jiang2024genai}. The labels in GenAI-Bench represent general user preferences rather than focusing on specific aesthetic concepts.

\textbf{VideoDPO} \citep{liu2025videodpo} is a preference dataset containing 10K synthesized video pairs. For each prompt, it uses the OmniScore metric \citep{liu2025videodpo}, which jointly evaluates visual fidelity and aesthetics, temporal smoothness across frames, and semantic alignment with the text, to select the best and worst samples from multiple generated videos, forming a preference pair.

As discussed in \cref{sec:exp-video}, we adopt different evaluation metrics for different types of reward models, and ultimately use the models' relative ranking as the final evaluation result \textbf{for the video-alignment capability of aesthetic reward models}. For AesRM-Base and AesRM-CoT, when computing the GSB score in \cref{eq:GSB}, we likewise perform bidirectional inference: we first feed, in order, the videos generated by the fine-tuned model and then by the pre-fine-tuned model to obtain an evaluation score $s_1$; we then reverse the order, feeding the videos generated by the pre-fine-tuned model followed by those from the fine-tuned model, to obtain a second evaluation score $s_2$. The score of the fine-tuned model relative to the pre-fine-tuned model is $(s_1 - s_2)/2$, which is then substituted into \cref{eq:GSB} to compute the GSB score. This bidirectional evaluation helps mitigate erroneous judgments caused by reward-model hallucinations, thereby improving the accuracy of the evaluation results.


\subsection{Aesthetic Baselines}
\phantomsection
\label{app:Baselines}

We further introduce the aesthetic baselines used in \cref{sec:exp-rm}. Specifically,

\textbf{LAP} \citep{schuhmann2022improved} is an aesthetic predictor built on CLIP ViT-L/14 \citep{radford2021learning} embeddings. It extracts an image’s CLIP feature vector and predicts its mean rating on the AVA dataset \citep{6247954} via a 5-layer MLP, which is used as the aesthetic score.

\textbf{ArtiMuse} \citep{cao2025artimuse} is an aesthetic reward model built on InternVL-3-8B \citep{zhu2025internvl3exploringadvancedtraining}. It introduces a Token-as-Score scheme that designates 101 predefined tokens as aesthetic rating tokens, corresponding to the integer scores from 0 to 100 annotated by human experts. The model is fine-tuned to predict these discrete tokens; during inference, the predicted token is mapped back to a numeric value, and the final aesthetic score is computed as the expected value over the probability distribution of all rating tokens.

\textbf{VADB} \citep{qiao2025vadb} is also built upon CLIP ViT-B/32 \citep{radford2021learning}, extending 2D convolution kernels to 3D to extract spatiotemporal video features. It adopts a dual-text-encoder design, including the original 12-layer CLIP Transformer for natural-language comments and an independent tag encoder for aesthetic tags such as symmetric composition and top lighting. Finally, similar to LAP, a lightweight MLP is applied on top of the CLIP embeddings to regress an aesthetic score.

\textbf{VideoAlign} \citep{liu2025improving} is built on Qwen2-VL-2B \citep{wang2024qwen2vlenhancingvisionlanguagemodels}. It inserts two context-agnostic tokens, \texttt{[VQ]} and \texttt{[MQ]}, immediately after the video and before the prompt tokens so that they attend only to visual content, and places a context-aware token, \texttt{[TA]}, after the full prompt tokens so it can attend to both the video and the text. The final-layer embeddings of these tokens are then mapped to the VQ, MQ, and TA scores via a shared linear layer.

\textbf{AesRM} differs from the above designs by framing aesthetic assessment as a pure next-token prediction task, thereby maximizing the VLM’s native capabilities \citep{wu2025rewarddance}. Leveraging expert aesthetic feedback in \cref{sec:data_collect} and a hierarchical aesthetic framework \cref{sec:Hierarchical}, AesRM-Base enables fast evaluation along three dimensions, VA, VF, and VP, while AesRM-CoT further provides chain-of-thought rationales over 15 fine-grained criteria, improving interpretability. This yields a high-quality and interpretable reward signal for video aesthetic alignment.

\subsection{Flow-RWR}
\phantomsection
\label{app:RWR}

Flow-RWR extends Reward-weighted Regression (RWR) \citep{peters2007reinforcement} from diffusion models \citep{lee2023aligningtexttoimagemodelsusing,furuta2024improvingdynamicobjectinteractions} to flow-based generative models, e.g., Wan2.2 trained with rectified flow. It can be viewed as a reward-weighted SFT algorithm. Formally, let $\mathbf{v}^*$ denote the velocity field of the target flow model and $\mathbf{v}_\theta(\mathbf{x}_t,t)$ the regressed velocity predicted by the generative model, where the noisy sample is constructed by linear interpolation
\begin{equation}
\mathbf{x}_t = (1-t)\,\mathbf{x}_0 + t\,\mathbf{x}_1,
\end{equation}
with $\mathbf{x}_0$ drawn from the real data distribution and $\mathbf{x}_1$ a noise sample. The Flow-RWR objective is a reward-weighted flow-matching loss:
\begin{equation}
\label{eq:flow-rwr}
\mathcal{L}_{\mathrm{RWR}}(\theta)
= \mathbb{E}\Big[\exp\big(R(\mathbf{x}_0,c)\big)\,\big\|\mathbf{v}^* - \mathbf{v}_\theta(\mathbf{x}_t,t,c)\big\|_2^2\Big].
\end{equation}
where $R(\cdot,\cdot)$ is the reward function and $c$ is the given condition, such as prompt.
In our setting, we use high-quality videos as references and employ AesRM to quantify the discrepancy between a model-generated video and its reference, using the resulting value to compute training weights. Specifically, for \textbf{AesRM-Base}, we sum the scores over the three dimensions (VA, VF, VP). For example, if the reference video is scored as $(1,1,1)$ relative to the fine-tuned model’s generated video on VA, VF, and VP, respectively, the weight in \cref{eq:flow-rwr} is $\exp(3)$, indicating a large gap and thus assigning a higher weight to this sample. For \textbf{AesRM-CoT}, we compute the weight by summing scores over 15 fine-grained criteria. For a sample $\mathbf{x}_i$, its Flow-RWR weight is
\begin{equation}
w_i = \exp\Big(\sum^{j=15}_{j=1} s_{i,j}\Big),
\end{equation}
where $s_{i,j}$ denotes the score for the $j$-th criterion.

In practice, we find that aggregating {bidirectional} AesRM comparisons yields a more stable reward signal \citep{wu2026editrewardhumanalignedrewardmodel}. Concretely, following the bidirectional evaluation protocol in \cref{app:Evaluation}, we first feed $(\mathbf{x}_{\text{gen}}, \mathbf{x}_{\text{ref}})$ into AesRM to obtain a score $s_1$, and then swap the order and evaluate $(\mathbf{x}_{\text{ref}}, \mathbf{x}_{\text{gen}})$ to obtain $s_2$. As the two evaluations should be anti-symmetric, we define the final reward as
\begin{equation}
s=\frac{s_1-s_2}{2},
\end{equation}
and compute the Flow-RWR weight as
\begin{equation}
w=\exp(-s).
\end{equation}

\subsection{Pref-GRPO}
\phantomsection
\label{app:Pref-GRPO}

\textbf{GRPO on Flow Matching.}
The typical GRPO \citep{guo2025deepseek} untilizes the group-relative advantage to update models. Consider a group of $G$ generated videos $\{\mathbf{x}_0^i\}_{i=1}^{G}$, the advantage of the $i$-th video is
\begin{equation}
\label{eq:advantage}
\hat{A}_t^i
=
\frac{
R(\mathbf{x}_0^i,c) - \mathrm{mean}\!\left(\{R(\mathbf{x}_0^j,c)\}_{j=1}^{G}\right)
}{
\mathrm{std}\!\left(\{R(\mathbf{x}_0^j,c)\}_{j=1}^{G}\right)
}.
\end{equation}
where $c$ is the condition, i.e., the generation prompt.

The model is updated by maximizing the following objective
\begin{equation}
\mathcal{L}_{\mathrm{Flow\text{-}GRPO}}(\theta)
= \mathbb{E}_{c,\{x^i\}}
\Bigl[f(r,\hat{A},\theta,\epsilon,\beta)\Bigr],
\end{equation}
where
\begin{equation}
\label{eq:kl}
f(r,\hat{A},\theta,\epsilon,\beta)
=
\frac{1}{G}\sum_{i=1}^{G}\frac{1}{T}\sum_{t=0}^{T-1}
\min\!\Bigl(
r_t^i(\theta)\hat{A}_t^i,\;
\mathrm{clip}\bigl(r_t^i(\theta),1-\epsilon,1+\epsilon\bigr)\hat{A}_t^i
\Bigr)
-\beta\,D_{\mathrm{KL}}(\pi_\theta\|\pi_{\mathrm{ref}}),
\end{equation}
with
\begin{equation}
r_t^i(\theta)
=
\frac{p_\theta\!\left(\mathbf{x}_{t-1}^i \mid \mathbf{x}_t^i, c\right)}
     {p_{\theta_{\mathrm{old}}}\!\left(\mathbf{x}_{t-1}^i \mid \mathbf{x}_t^i, c\right)}.
\end{equation}

To estimate the KL divergence, the deterministic Flow-ODE sampling process, \(\mathrm{d}\mathbf{x}_t = v_t\,\mathrm{d}t\), is converted into an equivalent SDE \citep{liu2025flowgrpotrainingflowmatching,xue2025dancegrpounleashinggrpovisual}:
\begin{equation}
\mathrm{d}\mathbf{x}_t
=
\Bigl(v_\theta(\mathbf{x}_t,t)+\frac{\sigma_t^2}{t}\bigl(\mathbf{x}_t+(1-t)\,v_\theta(\mathbf{x}_t,t)\bigr)\Bigr)\mathrm{d}t
+\sigma_t\,\mathrm{d}\mathbf{w}_t.
\end{equation}

where $\mathrm{d}\mathbf{w}_t$ denotes Wiener process increments and $\sigma_t$ controls the stochasticity. And via Euler-Maruyama discretization, we have
\begin{equation}
\label{eq:sde}
\mathbf{x}_{t+\Delta t}
= \mathbf{x}_t
+ \left(
\mathbf{v}_\theta(\mathbf{x}_t,t)
+ \frac{\sigma_t^2}{2t}\Bigl(\mathbf{x}_t + (1-t)\,\mathbf{v}_\theta(\mathbf{x}_t,t)\Bigr)
\right)\Delta t
+ \sigma_t \sqrt{\Delta t}\,\mathbf{\epsilon},
\quad \epsilon \sim \mathcal{N}(0,I).
\end{equation}
where $\sigma_t = a\sqrt{\frac{t}{1-t}}$ and $a$ is a scalar hyper-parameter that controls the noise level. Then, by \cref{eq:sde}, we can estimate the Gaussian transition distributions $\pi_{\theta}(\mathbf{x}_{t+\Delta t}\mid \mathbf{x}_{t},c)$ and $\pi_{\text{ref}}(\mathbf{x}_{t+\Delta t}\mid \mathbf{x}_{t},c)$, thereby obtaining an estimate of the KL term in \cref{eq:kl}.

\textbf{Pref-GRPO on Flow Matching.} Pref-GRPO \citep{wang2025pref} reformulates the optimization objective of GRPO as {pairwise preference fitting}. Specifically, rather than relying on absolute reward scores $R(\cdot,\cdot)$, Pref-GRPO evaluates relative preferences among generated samples, providing more stable and informative advantages for policy optimization while reducing susceptibility to reward hacking \citep{wang2025pref}. Concretely, Pref-GRPO performs pairwise comparisons within each group and uses each sample's win rate (i.e., the probability of being preferred over other samples in the same group) as the reward signal. This reward is then plugged into \cref{eq:advantage} for advantage estimation and the subsequent optimization procedure. Pref-GRPO naturally aligns with AesRM, which outputs pairwise preference judgments, and is therefore chosen as our RL-style post-training algorithm.

\subsection{More Quantitative Results}
\phantomsection
\label{app:More-Results}

In this section, we report additional quantitative results, including (i) the detailed accuracy of AesRM on AesVideo-Bench across the three dimensions VA, VF, and VP, and (ii) the video aesthetic alignment results achieved with Pref-GRPO.

\textbf{Detailed Accuracy on AesVideo-Bench.} \cref{sec:exp-rm} reports the binary accuracy of AesRM on AesVideo-Bench. However, AesVideo-Bench is a three-class task, where the model makes predictions along three dimensions: VA, VF, and VP. Therefore, \cref{tab:app_acc_aesrm_base} and \cref{tab:app_acc_aesrm_cot} further present more detailed results for AesRM-Base and AesRM-CoT, including binary accuracy, three-class accuracy (i.e., for each sample, the model must not only predict the preference correctly but also simultaneously predict the VA/VF/VP labels), average accuracy (the mean accuracy over VA, VF, and VP), as well as the per-dimension accuracies for VA, VF, and VP. We observe that our three-stage training strategy, together with self-consistency-based CoT synthesis and process rewards, consistently improves AesRM’s accuracy across all dimensions. 

\begin{table}[tb]
  \caption{AesRM-Base performance on AesVideo-Bench in terms of binary accuracy (Binary), three-class accuracy (Three-class), average accuracy (Avg.), and per-dimension accuracies for VA, VF, and VP. Our three-stage training strategy especially on binary and three-class accuracy.}
  \label{tab:app_acc_aesrm_base}
  \centering
  \begin{tabular}{@{}lcccccc@{}}
    \toprule
    {Model (\%)} & {Binary} & {Three-Class}& {Avg.} & {VA}  & {VF} & {VP} \\
    \midrule
    InternVL3.5                 & 59.67 & 16.08 & 51.95 & 55.12 & 58.00 & 42.74 \\
    + \textit{Cold-Start and GRPO}           & 66.61 & 25.34 & 64.42 &63.90 & {84.16} & 45.21 \\
    + \textit{Atomic Learning}  & \textbf{68.86} &\textbf{26.89} &\textbf{ 65.38}& \textbf{64.77}& \textbf{85.25}& \textbf{46.11} \\
    \bottomrule
  \end{tabular}
\end{table}
\begin{table}[tb]
  \caption{AesRM-CoT performance on AesVideo-Bench in terms of binary accuracy (Binary), three-class accuracy (Three-class), average accuracy (Avg.), and per-dimension accuracies for VA, VF, and VP. Our three-stage training strategy, together with ensemble-based CoT synthesis and process rewards, consistently improves performance, especially on binary and three-class accuracy.}
  \label{tab:app_acc_aesrm_cot}
  \centering
  \begin{tabular}{@{}lcccccc@{}}
    \toprule
    {Model (\%)} & {Binary} & {Three-Class} & {Avg.} & {VA} & {VF} & {VP} \\
    \midrule
    InternVL3.5                 & 59.67 & 16.08 & 51.95&55.12 & 58.00 & 42.74 \\
    + \textit{Cold-Start and GRPO}           & 65.51 & 24.95 & 64.87 &65.05 & \textbf{85.29} & \textbf{44.27} \\
    + \textit{Ensemble CoT}              & 66.61 & 25.31 & 64.97 &66.55 & {84.33} & 44.03 \\
    + \textit{Process Reward}     & 66.99 & 25.77 & \textbf{65.39} &\textbf{66.76} &84.70 & {44.72} \\
    + \textit{Atomic Learning}     & \textbf{69.41} & \textbf{29.84} & 64.88&{66.33} & 85.23 & {43.09} \\
    \bottomrule
  \end{tabular}
\end{table}

\textbf{Video Aesthetic Alignment based on Pref-GRPO.} Following the experimental setting in \cref{sec:exp-video}, we report post-training results for Wan2.2 using different reward models in \cref{tab:wan_grpo}, which show that during the RL phase, AesRM can still effectively improve various aspects of video aesthetics.

\begin{table}[tb]
  \captionsetup{font=small,skip=6pt}
  \caption{Video aesthetic evaluation of Wan2.2 fine-tuned with various aesthetic reward models based on Pref-GRPO.}
  \label{tab:wan_grpo}
  \centering
  \resizebox{\linewidth}{!}{%
  \begin{tabular}{@{}lccccccc@{}}
    \toprule
    \textbf{Model} & \textbf{LAP $\uparrow$} & \textbf{VideoAlign-VQ $\uparrow$} &
    \textbf{AesRM-Base $\uparrow$} & \textbf{AesRM-CoT $\uparrow$} & \textbf{UniRM \cite{wang2025unified-1} $\uparrow$} & \textbf{Avg. Rank $\downarrow$} \\
    \midrule
    Wan2.2-TI2V-5B            & 5.505& -0.592 & --     & --     & 2.817 & 5.00  \\
    + \textit{LAP} \cite{schuhmann2022improved}           & 5.523  & -0.535 & 16.073 & 33.258 & 2.845 & 2.67 \\
    + \textit{VideoAlign-VQ} \cite{liu2025improving}  & 5.529  & -0.534  & 11.148 & 31.865 & 2.855 & 3.17 \\
    \grayrow
    + \textit{AesRM-Base}     &  5.531  & -0.507 & 14.807 & 37.161 & 2.864 & 1.80\\
    \grayrow
    + \textit{AesRM-CoT}      & {5.536}  & -0.505 & 16.388 & 33.503 & 2.849 & 1.60 \\
    \bottomrule
  \end{tabular}%
  }
\end{table}


\subsection{More Qualitative Cases}
\phantomsection
\label{app:More-Genration-Cases}

\begin{figure}[t!]
\centering
\includegraphics[width=1.0\linewidth]{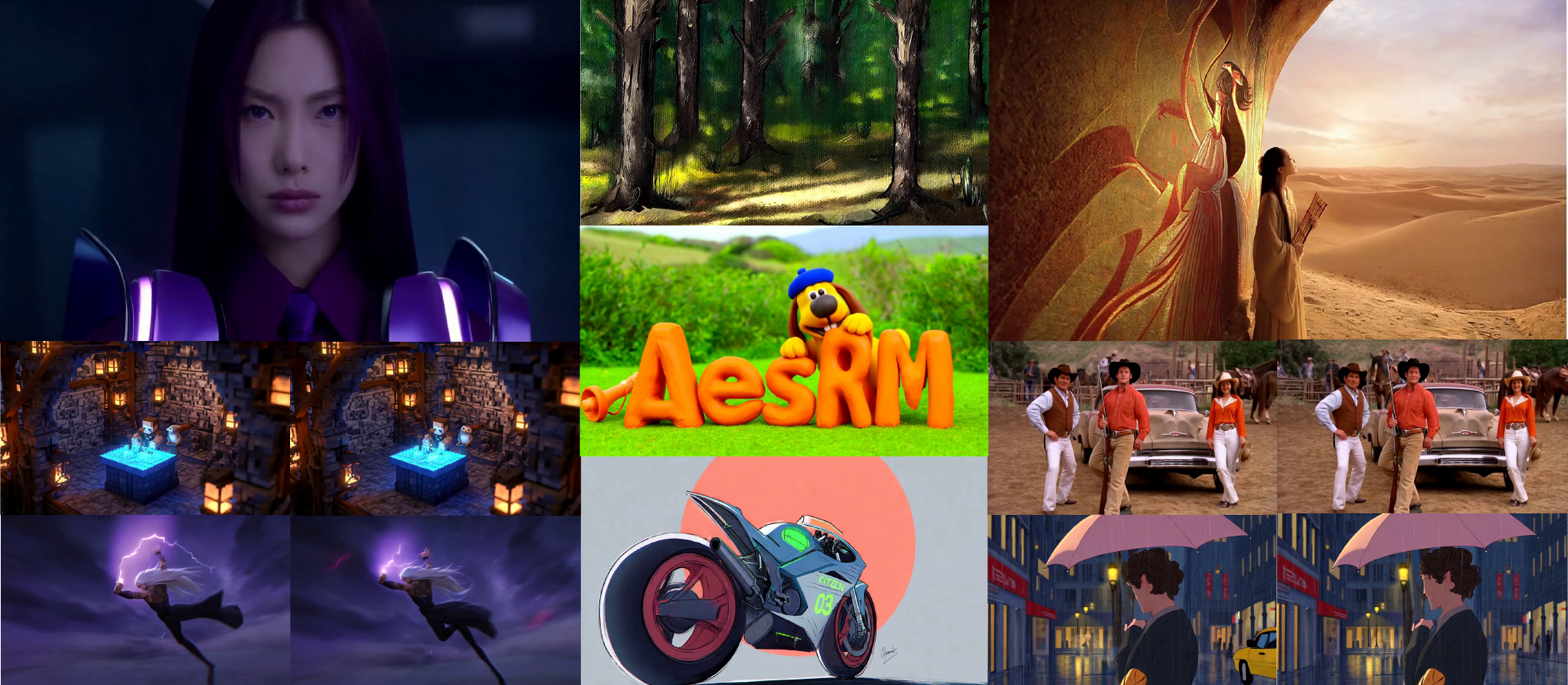}
\caption{A comparison between videos generated by Wan2.2-TI2V-5B \citep{wan2025wan} under the same prompts and those produced by Wan2.2-TI2V-5B with AesRM in \cref{fig:pipeline}. We observe that the original Wan2.2 performs worse in terms of fine-grained details, contrast, sharpness, and shot composition, often resulting in blurry faces, underexposed visuals, and less impactful subject framing. }
\label{fig:wan}
\end{figure}

In this section, we present additional qualitative examples to illustrate how AesRM improves video aesthetics. First, \cref{fig:wan} shows videos generated by the original, non-fine-tuned Wan2.2 under the same prompt as in \cref{fig:pipeline}. By comparison, we observe that the original Wan2.2 produces faces with strong smearing artifacts and poorly defined facial contours. The overall composition lacks visual impact, with limited detail richness and overly strong color contrast, resulting in a darkened and yellowish appearance. These issues are substantially alleviated after fine-tuning with AesRM.

\begin{figure}[t!]
\centering
  \includegraphics[width=1\linewidth]{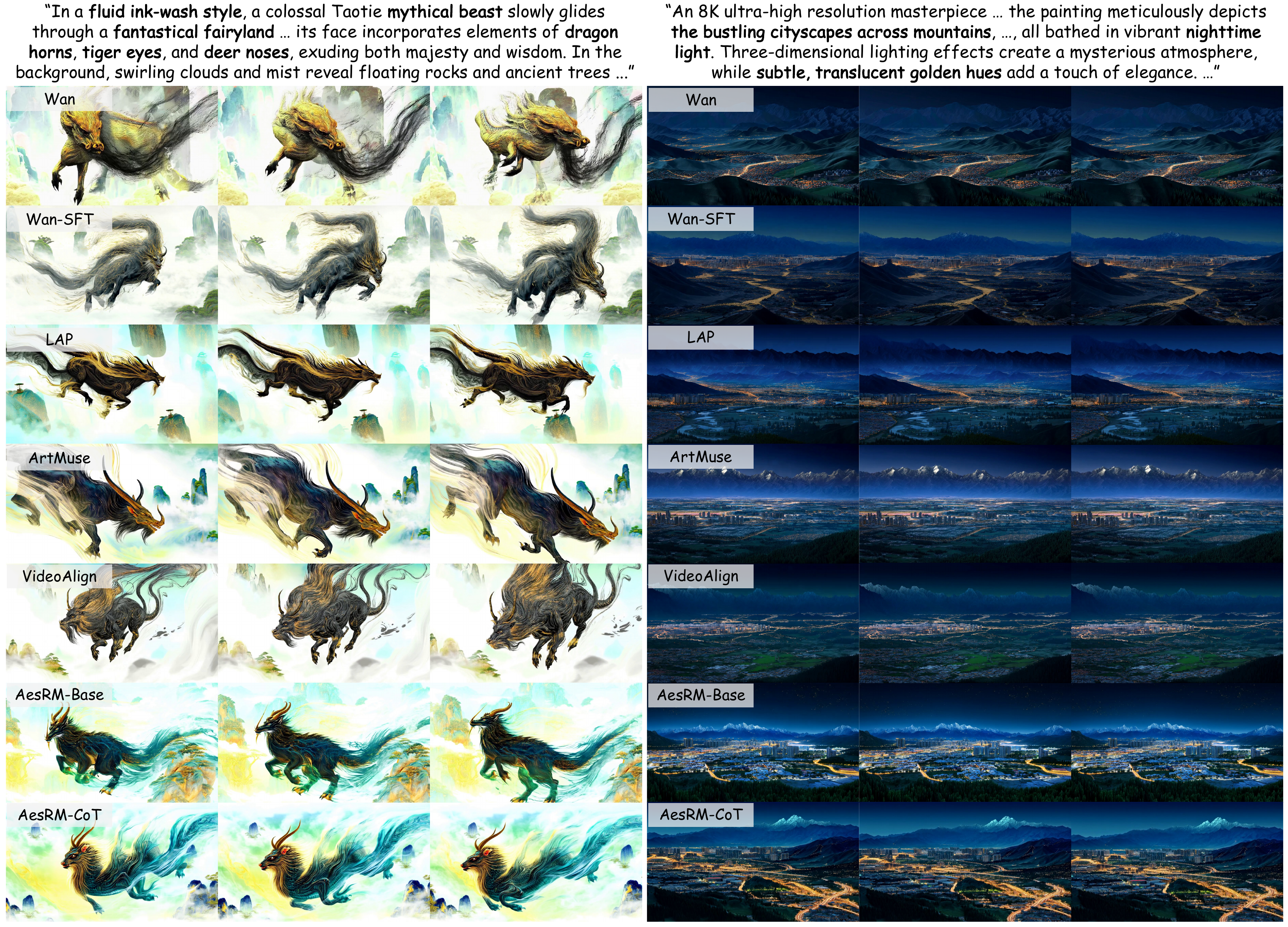}
\caption{More visualization results of Wan2.2 generations under different aesthetic reward models using Flow-RWR. Fine-tuning with AesRM-Base or AesRM-CoT yields better aesthetics. \textbf{Left:} Videos produced by the AesRM-Base and AesRM-CoT-fine-tuned models exhibit richer color and a more centered composition, presenting the subject more completely. \textbf{Right:} Videos generated by the AesRM-Base and AesRM-CoT–fine-tuned models show richer scene details and more vivid colors, avoiding frames that fall into near-complete darkness.}
\label{fig:more_vis_1}
\end{figure}

\begin{figure}[t!]
\centering
  \includegraphics[width=1\linewidth]{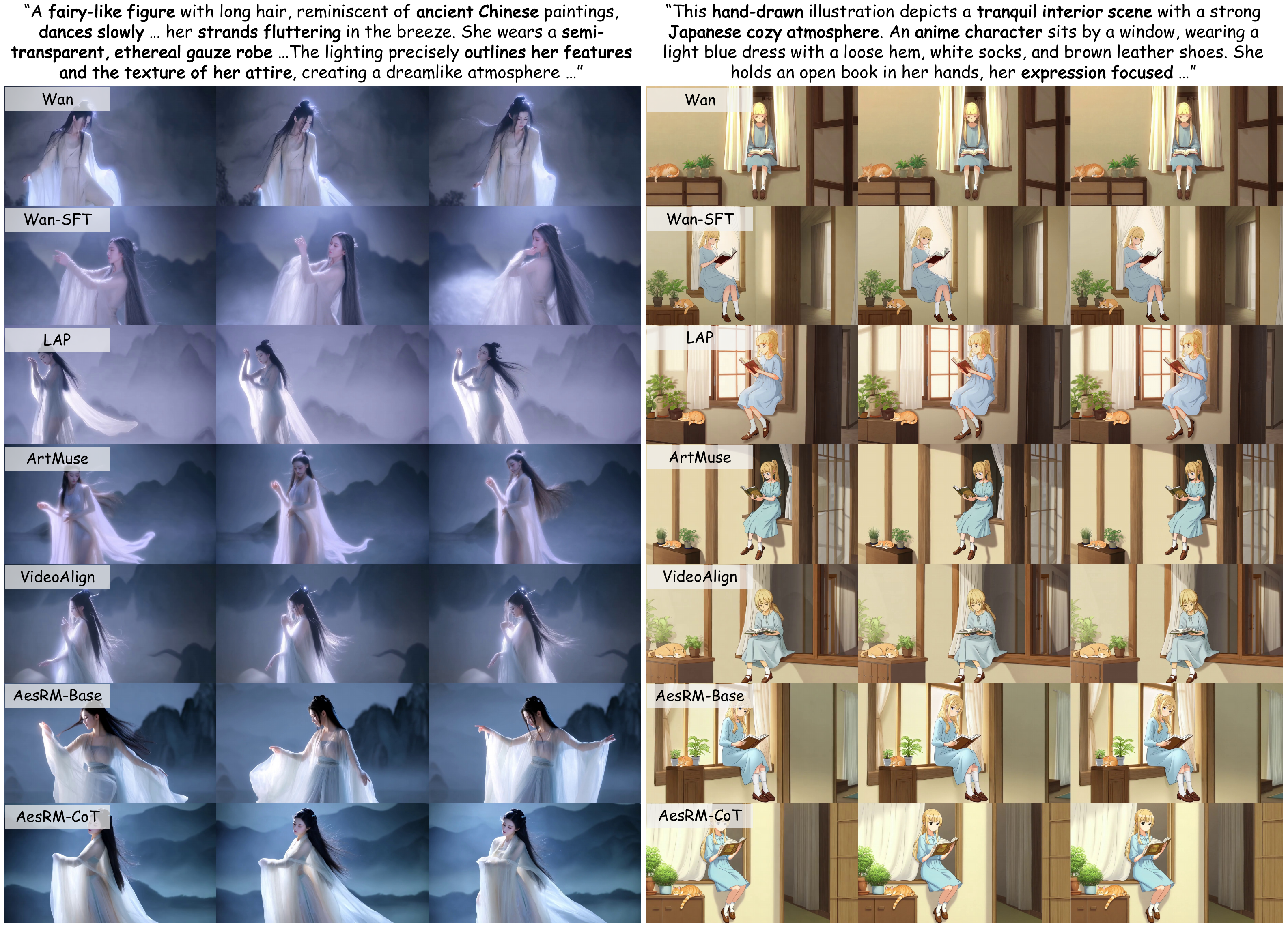}
\caption{More visualization results of Wan2.2 generations under different aesthetic reward models using Flow-RWR. Fine-tuning with AesRM-Base or AesRM-CoT yields better aesthetics. \textbf{Left:} Videos fine-tuned with AesRM exhibit natural color rendering without a noticeable pink tint, and the character is sharp with rich details (e.g., clear details in the woman’s clothing). \textbf{Right:} AesRM improves overall color quality and fine-grained details (e.g., facial features and book textures), while keeping the subject centered with a clear, well-structured composition.}
\label{fig:more_vis_2}
\end{figure}

Next, we present more cases of Wan2.2 fine-tuned with the Flow-RWR post-training strategy under different reward models, as shown in \cref{fig:more_vis_1} and \cref{fig:more_vis_2}. We observe that Wan2.2 fine-tuned with AesRM-Base and AesRM-CoT generates videos with richer details and better visual sharpness. For example, the puppy remains clearly defined even under mixed lighting, rather than appearing dark and blurry. 

\begin{figure}[t!]
\centering
  \includegraphics[width=1\linewidth]{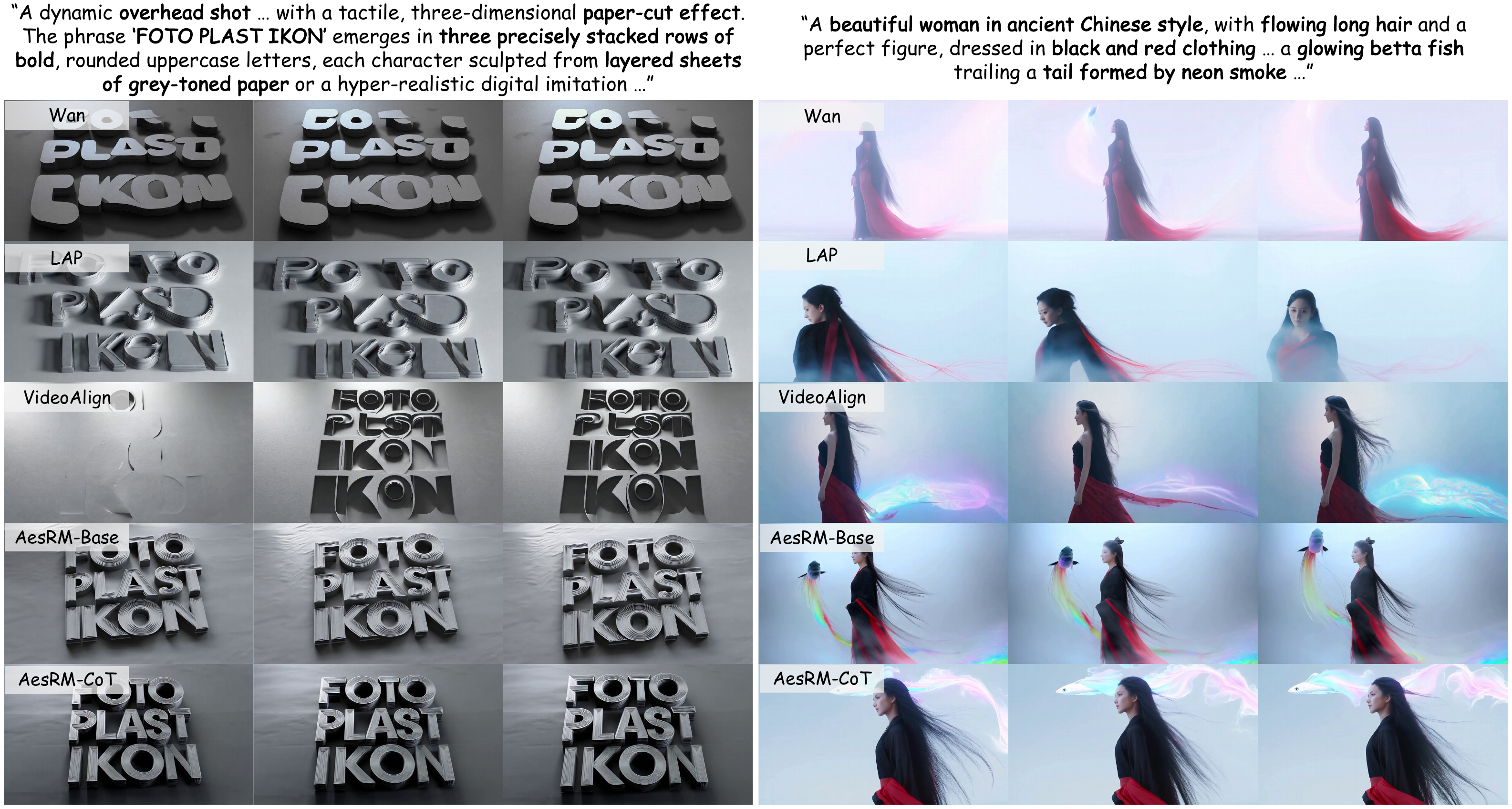}
\caption{More visualization results of Wan2.2 generations under different aesthetic reward models using Pref-GRPO. Fine-tuning with AesRM-Base or AesRM-CoT yields better aesthetics. \textbf{Left:} After fine-tuning with AesRM-Base or AesRM-CoT, the generated videos accurately render the target letters and satisfy the requirement of being “sculpted from layered sheets of grey-toned paper.” \textbf{Right:} AesRM helps the model generate key elements (e.g., the fish) more faithfully while improving overall sharpness and color richness.}
\label{fig:more_vis_GRPO}
\end{figure}

\end{document}